\documentclass[lettersize,journal]{IEEEtran}
\usepackage{amsmath,amsfonts}
\usepackage{algorithmic}
\usepackage{algorithm}
\usepackage{array}
\usepackage[caption=false,font=normalsize,labelfont=sf,textfont=sf]{subfig}
\usepackage{textcomp}
\usepackage{stfloats}
\usepackage{url}
\usepackage{verbatim}
\usepackage{graphicx}
\usepackage{cite}
\usepackage{amsmath,amsfonts,bm}

\def\eqref#1{equation~\ref{#1}}
\def\1{\bm{1}}

\def\vdelta{{\bm{\delta}}}
\def\va{{\bm{a}}}
\def\vb{{\bm{b}}}

\def\vf{{\bm{f}}}

\def\vk{{\bm{k}}}

\def\vx{{\bm{x}}}
\def\vy{{\bm{y}}}

\DeclareMathAlphabet{\mathsfit}{\encodingdefault}{\sfdefault}{m}{sl}
\SetMathAlphabet{\mathsfit}{bold}{\encodingdefault}{\sfdefault}{bx}{n}

\DeclareMathOperator{\sign}{sign}

\usepackage{hyperref}
\usepackage{url}

\usepackage{epsfig}
\usepackage{graphicx}
\usepackage{amsmath}
\usepackage{amssymb}
\usepackage{url}
\usepackage{multirow}
\usepackage{algorithm}
\usepackage{algorithmic}
\usepackage{booktabs}
\usepackage{ragged2e}
 \usepackage{makecell, rotating}
\usepackage{comment}
\usepackage{enumitem}

\usepackage{xcolor}
\usepackage{colortbl}
\definecolor{mygray}{gray}{.9}
\definecolor{mypink}{rgb}{.99,.91,.95}
\definecolor{mycyan}{cmyk}{.3,0,0,0}

\newcolumntype{g}{>{\columncolor{mygray!50}}c}
\newcolumntype{a}{>{\columncolor{mygray}}r}
\def\ie{\emph{i.e.}}
\def\eg{\emph{e.g.}}

\begin{document}

\title{Revisiting and Advancing Adversarial Training Through A Simple Baseline}

\author{Hong Liu  
        % <-this % stops a space
\thanks{H. Liu is with the National Institute of Informatics, Tokyo, 101-8430, Japan. (e-mail: lynnliu.xmu@gmail.com)}% <-this % stops a space
\thanks{Our code will be released based on the acceptance.}
\thanks{Manuscript received November 2023.}}



\maketitle

\begin{abstract}
In this paper, we delve into the essential components of adversarial training which is a pioneering defense technique against adversarial attacks. 
We indicate that some factors such as the loss function, learning rate scheduler, and data augmentation, which are independent of the model architecture, will influence adversarial robustness and generalization.
When these factors are controlled for, we introduce a simple baseline approach, termed \textit{SimpleAT}, that performs competitively with recent methods and mitigates robust overfitting.
We conduct extensive experiments on CIFAR-10/100 and Tiny-ImageNet, which validate the robustness of SimpleAT against state-of-the-art adversarial attackers such as AutoAttack.  
Our results also demonstrate that SimpleAT exhibits good performance in the presence of various image corruptions, such as those found in the CIFAR-10-C.
In addition, we empirically show that SimpleAT is capable of reducing the variance in model predictions, which is considered the primary contributor to robust overfitting. 
Our results also reveal the connections between SimpleAT and many advanced state-of-the-art adversarial defense methods.
\end{abstract}


\section{Introduction}
\IEEEPARstart{R}{ecent} research \cite{li2023trustworthy} has shown that adversarial examples or out-of-distribution examples pose a serious threat to the robustness of deep models, leading to significant security issues.
Consequently, there is an increasing demand for algorithms that can improve the robustness and generalization of these models.
To this end, various methods have been proposed to defend against different types of adversarial attacks \cite{miller2020adversarial,ortiz2021optimism}.

In a nutshell, they can be categorized into various categories, including data augmentation \cite{alayrac2019labels,carmon2019unlabeled,sehwag2021improving,rebuffi2021fixing,rebuffi2021data,li2023data,Wang2023BetterDM}, architecture designing \cite{xie2019feature,xiao2019enhancing,chen2020anti,wu2021wider}, input transformation \cite{yang2019menet,dong2020api,mao2021adversarial,yoon2021adversarial}, certified defenses \cite{pmlr-v97-cohen19c,salman2020denoised,carlini2023certified}, and adversarial training \cite{madry2017towards,kannan2018adversarial,zhang2019theoretically,wang2019improving,zhang2021geometryaware,jorge2022make,Pang2022RobustnessAA,dong2022exploring,Liu2023MitigatingRO}.
Among them, adversarial training has been shown to be the most effective method by \cite{athalye2018obfuscated}, which involves training a robust model using min-max optimization \cite{madry2017towards,jorge2022make}.

The most widely used benchmark for comparing methods in terms of adversarial robustness has been CIFAR-10 \cite{croce2021robustbench,stutz2021relating}, where performance has steadily improved over the past few years.
However, the training schemes and objectives are quite different. 
Therefore, we perform a careful analysis and find that auxiliary factors, including loss function, learning rate scheduler, and data augmentation, have a significant impact on performance. 
When properly controlling these variations, the simple combination of these factors works surprisingly well, outperforming most state-of-the-art methods. 

Firstly, the training objective in previous works remains largely unchanged, with the cross-entropy loss as the default for adversarial training. 
Recent research \cite{hui2021evaluation,hu2022understanding} analyzes the reasons for the effectiveness of square loss in terms of robustness and calibration.
However, the observed improvements are relatively small when compared to various state-of-the-art methods for defending against adversarial attacks.
This raises a question: \textit{Can square loss be also effective for adversarial training as well?} 
Secondly, recent methods often adhere to a classical experimental setup, such as the one in \cite{rice2020overfitting}, where stochastic gradient descent with multiple step decays (MultiStepLR) is used.
However, directly substituting the cross-entropy loss with the squared loss in adversarial training could result in training issues, and all learning rate schedules are subject to robust overfitting \cite{stutz2021relating}. 
Therefore, this prompts another question: \textit{Are there better learning rate schedules for adversarial training?}
Thirdly, both \cite{li2023data} and \cite{Wang2023BetterDM} suggest that data augmentation plays a crucial role in improving adversarial robustness, but they do not delve into the specific aspects of effective data augmentation. 
So, \textit{is data augmentation the key to enhancing robustness, and if so, which techniques are the most impactful?}

To answer these, we conducted extensive experiments to study the effect of these factors and discovered several intriguing observations that offer novel perspectives for adversarial training.
Firstly, models trained with the square loss outperform those trained with cross-entropy loss and are equally competitive as models trained with cross-entropy loss together with erasing-based data augmentations \cite{devries2017improved,zhong2020random}. 
Secondly, we observe that combining a cyclic learning rate scheduler with either square loss or erasing-based data augmentation could mitigate robust overfitting while achieving comparable robust performance. 
These observations differ from previous works \cite{rice2020overfitting,pang2021bag}, where using each  technique separately does not reduce the risk of robust overfitting effectively.
Thirdly, deep models trained with the rescaled square loss achieve better results for both natural and adversarial accuracy, which is a sign of a good trade-off.
With the above observations, we introduce a simple baseline, called \textit{SimpleAT}, which remarkably outperforms various cutting-edge defense methods \cite{zhang2019theoretically,Liu2023MitigatingRO,Pang2022RobustnessAA,li2023data,Wang2023BetterDM}. 
SimpleAT highlights the significance of the rescaled square loss, cyclic learning rate scheduler, and erasing-based data augmentations in enhancing model robustness and generalization.

Generally speaking, no matter whether under PGD-based adversarial training \cite{madry2017towards} or FGSM-based adversarial training \cite{jorge2022make},  SimpleAT achieves state-of-the-art results on CIFAR-10/100 and Tiny-ImageNet. 
For example, SimpleAT achieves about 52.3\% adversarial accuracy and 85.3\% natural accuracy on CIFAR-10 without using extra synthesized unlabelled data, which demonstrate competitive results when compared to those reported in recent state-of-the-art methods \cite{rebuffi2021data,li2023data}.
Moreover, we conduct experiments on corruption datasets such as CIFAR-10-C, where the model trained with SimpleAT can produce higher performance (with 2.7\% accuracy gain) compared to that trained with the default settings in \cite{kireev2022effectiveness}. 

We further explain the reason why the SimpleAT method is effective from two different theoretical perspectives: \textit{bias-variance theory and logit penalty}.
Firstly, we utilize the adversarial bias-variance decomposition technique \cite{yu2021understanding} to understand the robust generalization of the SimpleAT method.  
We observe a consistent phenomenon that the variance of the model trained using SimpleAT is extremely low, which indicates that SimpleAT can make adversarial training more stable. 
In other words, SimpleAT mainly reduces adversarial bias, which helps to improve the generalization.
Secondly, we compare our SimpleAT with recent logit penalty methods, and we find that SimpleAT tends to reduce confidence in the true class. 
This is because overconfidence in adversarial examples can easily lead to robust overfitting \cite{Liu2023MitigatingRO} and worse generalization \cite{stutz2020confidence}. 

This paper is organized as follows:
Section \ref{sec2} introduces the basic techniques we used that form the foundation of this work.
From Section \ref{sec3} to Section \ref{sec6}, we  provide a comprehensive comparison and evaluation of related works alongside ours.
Section \ref{sec7} explains the reason why square loss makes sense from two theoretical perspectives.
At last,  Section \ref{sec8} concludes this paper.

\section{Methodology}
\label{sec2}

\subsection{Adversarial Training Baseline} \label{sec21}

The adversarial training (AT) paradigm underpins the majority of recently developed reliable mechanisms \cite{madry2017towards,rice2020overfitting,jorge2022make}.
In this paradigm, given a CNN model $\vf_{\theta}(\cdot)$ with parameter $\theta$, an input image $\vx$ with its ground-truth label $y$, AT is usually formulated as following \textit{min-max optimization}:
\begin{align}\label{eq1}
\min\nolimits_{\theta} \max\nolimits_{\vdelta} \mathcal{L}(\vf_{\theta}(\vx+\vdelta), y),  
\end{align}
where the outer optimization is to learn the parameter $\theta$ by minimizing the training objective $\mathcal{L}$, and the inner optimization is to generate the adversarial perturbation $\vdelta$ by maximizing the loss function $\mathcal{L}$.
Generally speaking, the training objective $\mathcal{L}$ is often the cross-entropy loss. 
There are two main learning paradigms: PGD-AT \cite{madry2017towards} and FGSM-AT \cite{jorge2022make}.
Both use stochastic gradient descent to update model parameters during the outer minimization optimization when input with adversarial examples. 
The main distinction between them lies in the choice of the inner maximization method for generating the adversarial perturbation $\vdelta$.

\subsubsection{PGD-AT}
Especially, {PGD-AT} uses the projected gradient descent-based adversarial attack (called PGD-attack \cite{madry2017towards}), which iteratively updates the perturbation from a random Gaussian noise $\vdelta_0$. The formulation of a PGD attack can be shown as:
\begin{align}
	\vdelta_{i+1} &= \vdelta_i + \alpha\times \sign(\nabla\mathcal{L}(\vf_\theta(\vx+\vdelta),y))), \label{eq2} \\
	\vdelta_{i+1} &= \max\{\min\{ \vdelta_{i+1}, \epsilon \}, -\epsilon \}, \label{eq3}
\end{align}
where $\sign$ is the sign function, $\nabla\mathcal{L}$ is the gradient of target function \emph{w.r.t.} input data, $\alpha$ is the step-size, and $\epsilon$ is the pre-defined perturbation budget.

\subsubsection{FGSM-AT}
However, PGD-attack is inefficient for training due to its computational cost, which increases linearly with the number of steps.
To address this issue, a recent promising solution is FGSM-AT \cite{goodfellow2014explaining,Wong2020Fast}, which approximates the inner maximization in Eq. (\ref{eq1}) through single-step optimization, also known as FGSM-attack.
The FGSM-attack calculates the adversarial perturbation $\vdelta$ along the direction of the sign of the gradient, using the general formulation:
\begin{align}
	\vdelta &= \eta+\alpha\cdot\sign(\nabla\mathcal{L}(\vf_\theta(\vx+\eta),y)), \label{eq4}\\
	\vdelta &= \max\{\min\{ \vdelta, \epsilon \}, -\epsilon \}, \label{eq5}
\end{align}
where $\eta$ is drawn from a random uniform distribution within interval $[-\epsilon,\epsilon]$.
However, FGSM-AT methods usually suffer from catastrophic overfitting, in which a model becomes suddenly vulnerable to multi-step attacks (like PGD-attack).
To this end, \cite{jorge2022make} introduces a Noise-FGSM method that learns a prior noise around the clean sample without clipping process:
\begin{align}
	\hat{\vx} &= \vx + \eta,\\
	\vx_{adv} &= \hat{\vx} + \alpha\cdot\sign(\nabla\mathcal{L}(\vf_\theta(\hat{\vx}),y)) = \vx + \vdelta,
\end{align}
where 
$\hat{\vx}$ is the data augmented with additive noise, and $\vx_{adv}$ is the generated adversarial example for outer minimization optimization. 
This method is further named NFGSM-AT.

\begin{table*}[!tbph]
\centering
       \footnotesize
       \setlength\tabcolsep{10pt}
       \caption{Summary of various training protocols. \label{SVP}}
\begin{tabular}{cccccccc}
Method & Arch.   & Data Aug. & Loss Function                & LR Scheduler & Regularization & KD       & WA       \\ \midrule[1pt]
Baseline & ResNet-18      & random flips and cropping  & cross-entropy        & MultiStepLR  & $\times$       & $\times$ & $\times$ \\ %\hline
TRADES  & ResNet-18 & random flips and cropping  & cross-entropy        & MultiStepLR  & TRADES         & $\times$ & $\times$ \\ %\hline
IDBH   & ResNet-18  & IDBH (with RE)      & cross-entropy        & Cyclic LR    & $\times$       & $\times$ & $\times$ \\ %\hline
SCORES  & ResNet-18 & random flips and cropping  & square loss          & Cyclic LR    & TRADES         & $\times$ & $\times$  \\ %\hline
SRC   & ResNet-18 & random flips and cropping  & cross-entropy        & MultiStepLR  & SRC         & $\times$  & $\times$  \\ %\hline
KD-SWA  & ResNet-18 & random flips and cropping  & cross-entropy        & MultiStepLR  & Tf-KD         & $\surd$  & $\surd$  \\ %\hline
SimpleAT& ResNet-18 & IDBH (with RE)   & rescaled square loss & Cyclic LR    & $\times$       & $\times$ & $\surd$  \\ %\hline
\end{tabular}
\end{table*}

\subsubsection{Basic Setup}
On CIFAR-10/100, the adversarial training (AT) baseline uses ResNet-18 as the backbone.
Following the default setting in \cite{rice2020overfitting,stutz2021relating}, in our preliminary experiments, AT baseline trains the models for $200$ epochs and batch size 128. 
We use a stochastic gradient descent (SGD) optimizer to update the parameters of the trained model. We set the initial learning rate to $0.1$ which is reduced by the factor $0.1$ at both $100$-th and $150$-th epoch, which is the so-called MultiStepLR. 
We set a weight decay of $5\times 10^{-4}$ and a momentum of $0.9$.
We use random flips and cropping as the default data augmentation.
The training objective $\mathcal{L}$ is the default cross-entropy loss without using extra regularization. 
During model training, we use either PGD-attack or FGSM-attack method with the same setting to generate the adversarial perturbation $\vdelta$.
For example, we use $10$ iterations for PGD-attack with $\alpha =2/255$ and $\epsilon=8/255$ under $L_\infty$ adversarial constraint. 
Following \cite{simpleview}, we also refer to this basic setup as the training protocol. 
This training protocol is summarized in Table \ref{SVP}, which we call it the \textit{baseline training protocol}.

\subsection{Variations in Existing Training Protocols} \label{sec22}

In this paper, we study the essential factors in the AT baseline, that affect the performance.
We identify three modifications to the default training protocol that are essential, which include changes to the loss function, learning rate schedule, and data augmentation. 
These will be discussed in detail below.

\subsubsection{Loss Function}
It is important to select an appropriate training objective to optimize Eq. (\ref{eq1}).
The cross-entropy loss is commonly used, however, recent research \cite{Pang2022RobustnessAA} has shown that a robust classifier trained with this loss may encounter issues such as robust overfitting and trade-offs. 
Therefore, we should consider carefully when choosing a training objective.

A recent study by \cite{hui2021evaluation} finds that training with the square loss produces comparable or even better results than cross-entropy loss for various classification tasks in a range of datasets and architectures. 
Moreover, subsequent theoretical works \cite{hu2022understanding,zhou2022optimization} explore the intriguing properties of models trained using square loss, such as robustness, generalization, neural collapse, and global landscape. 
These investigations employ innovative techniques like neural tangent kernel \cite{hu2022understanding} or unconstrained feature models \cite{zhou2022optimization}.
However, these works mainly focus on standard image classification tasks where inputs are clean (natural) images.
Although some of them \cite{Pang2022RobustnessAA,hu2022understanding} have shown that using square loss helps achieve good robustness against adversarial examples, the performance gain is not satisfactory while there is still a significant gap to top-performing methods on RobustBench \cite{croce2021robustbench}.

In our study, we propose to use the rescaled square loss function (shortened as \emph{RSL}) that helps achieve superior performance, which is shown as follows:
\begin{align}
\label{eq8}
L(\vx,\vy) &= \|\vk\cdot(\vf(\vx)-M\cdot \vy_{hot})\|^2_2 \nonumber \\
           &= k\cdot (\vf_{y}(\vx)-M)^2+ \scalebox{1.0}{\(\sum\)}_{j=1,j\neq y}^{C}\vf_{j}(\vx)^2.
\end{align}
Here, we define the logit output of a deep model as $\vf$, which consists of $C$ components $\vf_i$ (where $i=1,2,...,C$) and $C$ represents the number of classes in a given dataset. 
The label vector $\vy_{hot}$ is encoded using one-hot encoding.
The parameter $\vk$ in Eq. (\ref{eq8}) controls the loss value at the true label index $y$ and the other parameter $M$ rescales the one-hot  vectors.

\subsubsection{Learning rate (LR) schedule}

In \cite{rice2020overfitting}, Rice \emph{et al.} conducts empirical studies on the effect of various learning rate schedules, including piecewise decay, multiple decays, linear decay, cyclic scheduler, and cosine scheduler. 
They find that using a piecewise decay LR scheduler with an early-stopping technique consistently can achieve good results.
Furthermore, \cite{pang2021bag} verifies these results. 
In addition, Stutz \emph{et al.} \cite{stutz2021relating} claim that learning rate schedules play an important role in how and when robust overfitting occurs. 
On the other hand, Wong \emph{et al.} \cite{Wong2020Fast} find that one cyclic learning rate can work well under the FGSM-AT, where the learning rate linearly increases from zero to a maximum learning rate and back down to zero \cite{smith2017cyclical}.
This one cyclic learning rate is also used in many following FGSM-AT schemes such as NFGSM-AT.
That is, PGD-AT and FGSM-AT commonly adopt very different training protocols.
On the other hand, some other works \cite{huang2020self,rebuffi2021data,Pang2022RobustnessAA,Wang2023BetterDM} also use a cosine LR scheduler to achieve good results. 
However, their framework works on learning a robust classifier by using semi-supervised learning with an extra synthesized dataset. 

This leads to the fact that it is difficult to determine that the actual performance improvement comes from a change in the learning rate because data augmentation can also mitigate robust overfitting while achieving performance gain \cite{stutz2021relating,li2023data}. 
Generally speaking, we observe that the tendency of recent studies is to use a cyclic-based scheduler\footnote{We think the cosine scheduler also belongs to the cyclic learning rates.}, because it helps find a flatter minimum in the robust loss landscape \cite{stutz2021relating}. 
Therefore, in this study, we follow this observation and use one cyclic LR scheduler as our default scheduler.
Our experiments show that combining the cyclic LR scheduler with the other two modifications not only mitigates robust overfitting but also achieves comparable robust performance. 

\subsubsection{Data Augmentation}
Data augmentation is a basic machine-learning technique that can enhance the accuracy and robustness of models. 
However, it may not be effective in adversarial training as robust overfitting tends to occur, as noted by \cite{rice2020overfitting}.
Although previous works \cite{rebuffi2021data,li2023data,Liu2023MitigatingRO} have used data augmentation to mitigate the risk of robust overfitting, in these methods, data regularization is often coupled with other factors and cannot clearly demonstrate its effectiveness.
Interestingly, we observe that these methods generally adopt the same data regularization strategy, namely random erasing\footnote{Random erasing \cite{zhong2020random} and CutOut \cite{devries2017improved} are two almost identical data regularization techniques that are proposed during the same period. They differ slightly in their technical implementation, but their basic idea is the same, which is to remove some parts of an image. Therefore, this article refers to them collectively as random erasing.} \cite{zhong2020random,devries2017improved}.
Therefore, we suppose that random erasing is the key to enhancing robustness.
In the following section, we use experiments to support this claim.

\subsubsection{Our New Training Protocol}
Based on the above analysis and observation, we define our training protocol as follows.
The setup of our protocol is essentially the same as the baseline setup, except for the training objective, learning rate scheduler, and data regularization.
In particular, we use the rescaled square loss as our training objective. 
We use the \textit{one cycle learning rate}, and we set the maximum learning rate to $0.2$ and the minimum learning rate to $0.0$. Unless otherwise stated, we set the number of the training epochs as $200$.
We use the IBDH data augmentation technique \cite{li2023data}, which includes the \textit{random erasing technique}.
Moreover, in our training protocol, we use the weight averaging technique (WA) to further enhance robustness and accuracy.
In Table \ref{SVP}, we present the default setup of our training protocol which is referred to as \textbf{SimpleAT}.

{SimpleAT} is a unified training protocol for both PGD-AT and FGSM-AT. 
For PGD-AT, we use a $10$-step PGD-attack to generate the adversarial examples.
We use the NFGSM method to generate adversarial samples for FGSM-AT.
In the next section, our experimental results show that SimpleAT can maintain a favorable equilibrium between robustness and accuracy, as well as reduce the risks of both robust overfitting and catastrophic overfitting.

% TODO: write Experiments
\section{Main Results\label{sec3}}

%\subsection{Experimental Settings}

The above section describes the training protocols of both baseline and our SimpleAT. 
We introduce some other settings.  

\textbf{Dataset and Network Setup.}
We conduct experiments on three widely-used datasets, \emph{i.e.,} CIFAR-10, CIFAR-100, and Tiny-ImageNet.
We also evaluate the model's robustness on CIFAR-10-C which includes 15 types of corruption categorized into four groups: noise, blur, weather, and digital \cite{hendrycks2019benchmarking} corruptions.
In most experiments, we use the ResNet-18 \cite{he2016deep} as the backbone.
Moreover, we use WideResNet \cite{BMVC2016_87} as the backbone to evaluate the robust performance of CIFAR-10.

\textbf{Evaluation Protocol.}
During the testing phase, we evaluate the accuracies on natural and adversarial images. 
To generate the adversarial images, we primarily use four attack methods: FGSM-attack, PGD-attack, C\&W-attack, and AutoAttack \cite{croce2020reliable}. 
%It is worth noting that AutoAttack is a strong and reliable ensemble of diverse parameter-free attacks, which is widely used to verify robustness.
We use common experimental settings for most experiments, where the perturbation budget is $\epsilon=8/255$.  
For the PGD-attack and C\&W-attack, we initialize the perturbation via a random Gaussian noise and then calculate the perturbation using  10-step iterations with $\alpha=2/255$. 
For the FGSM-attack, we use the original FGSM-attack (as Eq. (\ref{eq4}) and Eq. (\ref{eq5}) shown) with $\alpha=10/255$ and $\epsilon=8/255$.
For the AutoAttack, we follow the same setting in \cite{pang2021bag}, which is composed of an ensemble of diverse attacks, including APGD-CE \cite{croce2020reliable}, APGD-DLR \cite{croce2020reliable}, FAB attack \cite{croce2020minimally}, and Square attack \cite{andriushchenko2020square}.

% \subsection{}

We ablate our SimpleAT through controlled experiments.
Several intriguing properties are observed.

\subsection{Eﬀectiveness of Square Loss}

\begin{figure*}[!t]
    \centering
    \small
    \setlength\tabcolsep{-1pt}
    \begin{tabular}{ccc}
             (a) \makecell[c]{\includegraphics[width=0.32\linewidth]{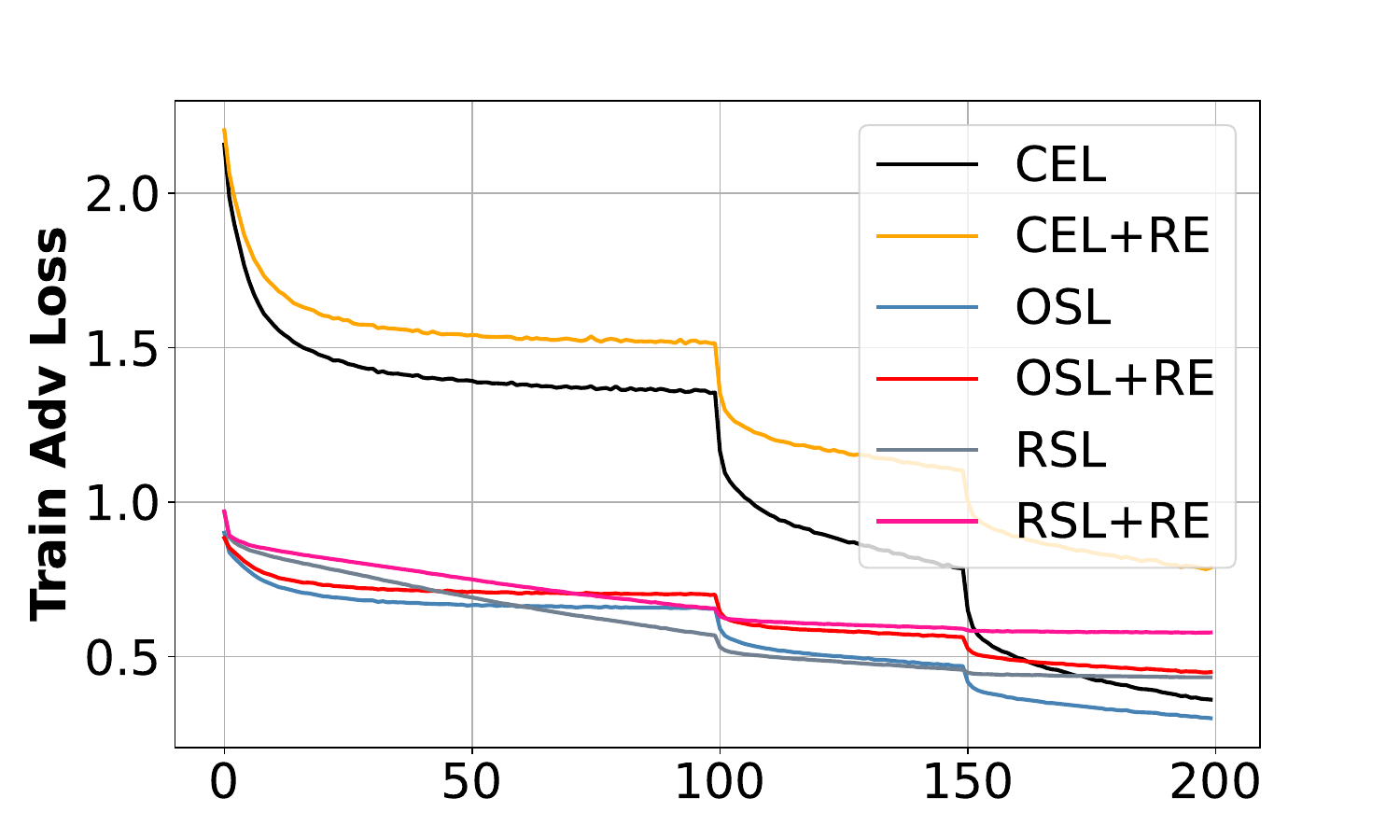}} & (b) \makecell[c]{\includegraphics[width=0.32\linewidth]{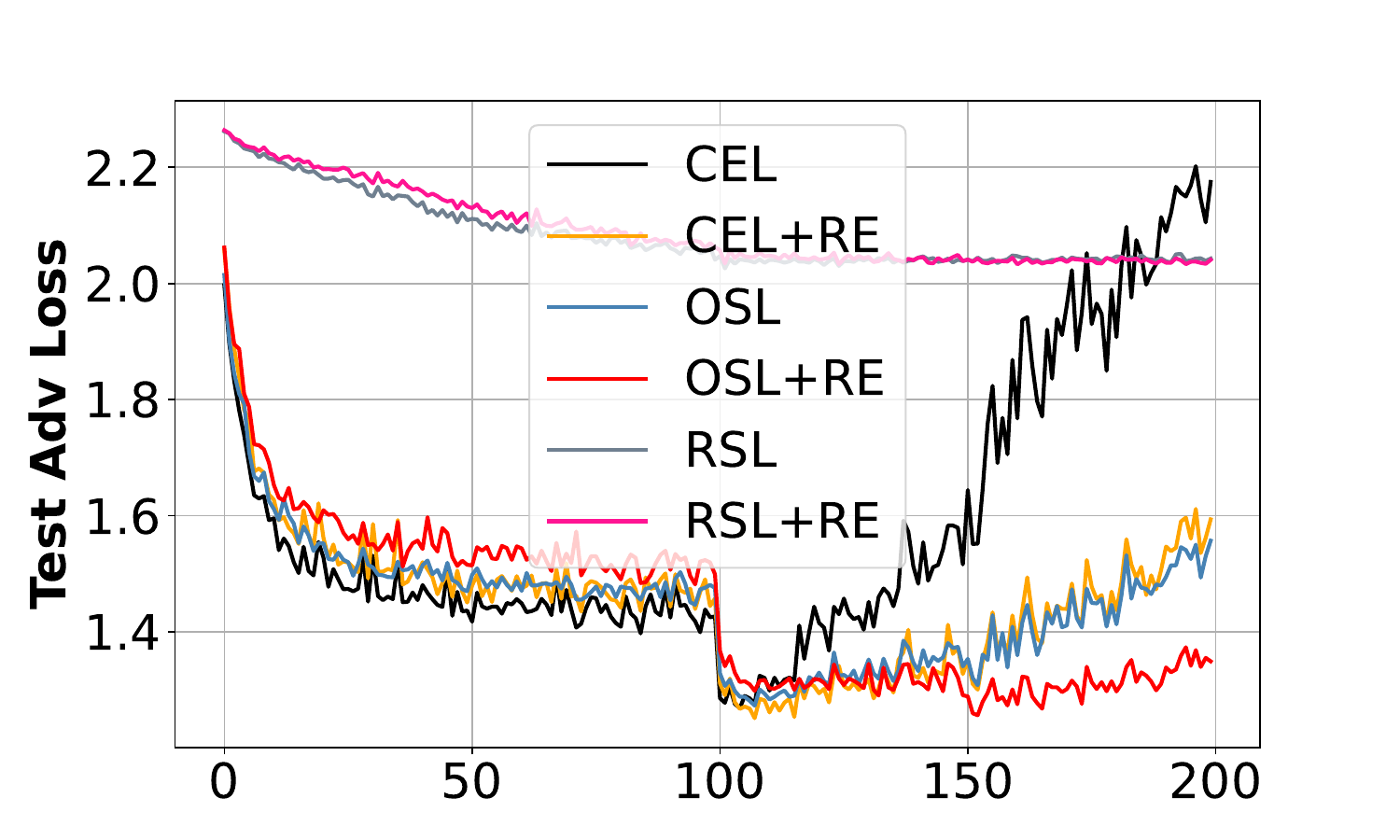}} & (c)   
             \makecell[c]{ \includegraphics[width=0.32\linewidth]{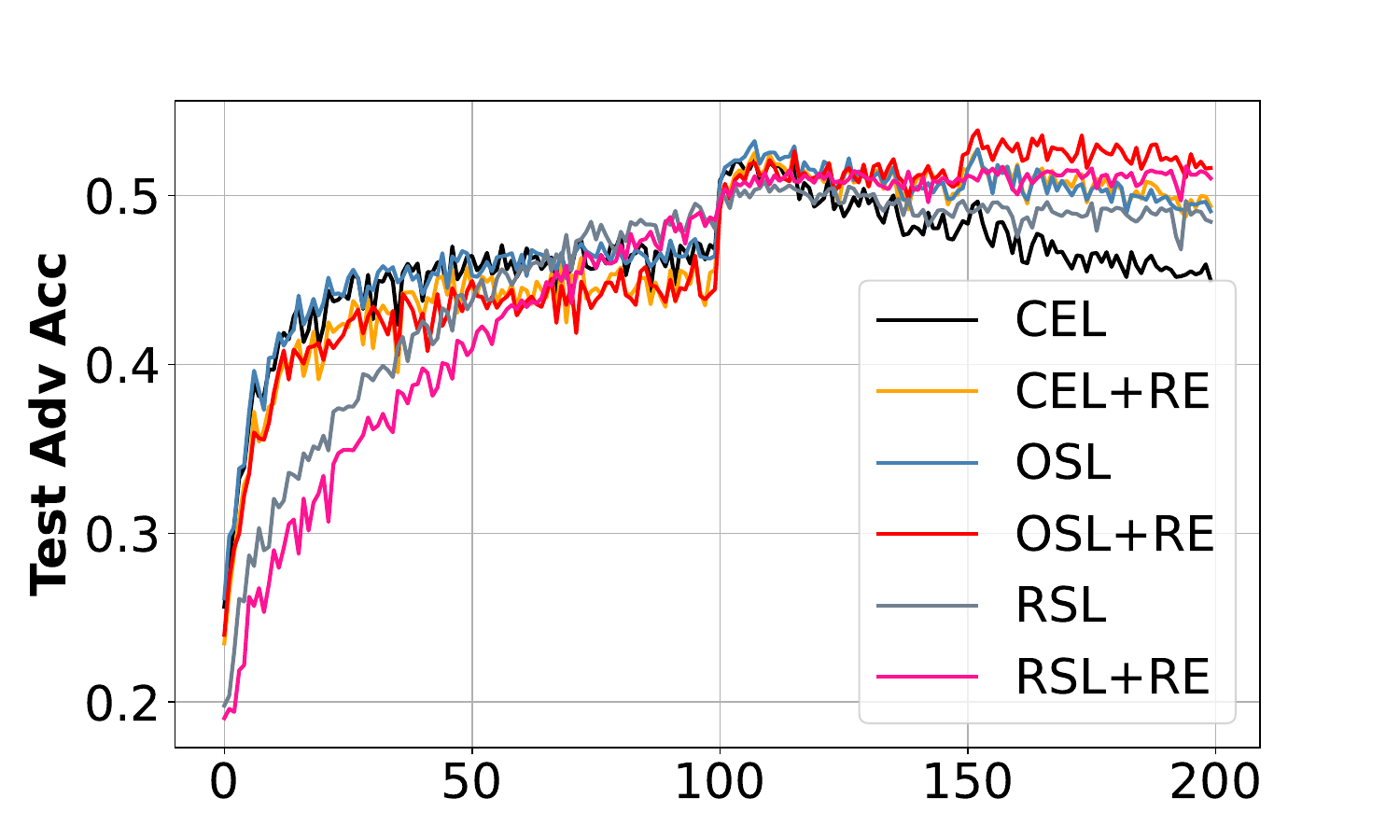}}
          \end{tabular} %\vspace{-0.5em}
    \caption{Analysis of square loss under the baseline training protocol. All results are evaluated on a robust ResNet-18 trained on CIFAR-10. Subfigure (a) shows the adversarial training loss curve, and subfigures (b) and (c) display the test learning curves.  \label{fig1}
    }
\end{figure*}

\begin{table}
       \centering
       \small
       \setlength\tabcolsep{3pt}
        \caption{\label{tab2} Quantitive results ($\%$) on CIFAR-10 by using CEL, OSL, and RSL. The ``Nat-D'' means the difference between the highest and final checkpoints for clean accuracy. ``PGD-D'' indicates the accuracy difference between the highest and final checkpoints under PGD-attack. ``AA-B'' reports the results of the best checkpoint under AutoAttack, ``AA-L'' is the results of the final checkpoint under AutoAttack, and ``AA-D'' is the difference between the highest and final checkpoints under AutoAttack. ``+RE'' means the model trained with random erasing. 
        }
       \scalebox{1.0}[1.0]{
      \begin{tabular}[b]{lccccc}
& Nat-D & PGD-D & AA-B & AA-L & AA-D  \\ \midrule[1pt]
CEL    & 2.52 & 7.14 & 47.88 & 41.07 & 6.81  \\
OSL   & 4.40 & 4.13 & 48.43 & 44.73 & 3.70  \\
RSL   & 1.60 & 2.59 & 45.49 & 41.97 & 3.52  \\\midrule[0.5pt]
CEL+RE & 3.18 & 3.45 & 47.55 & 45.52 & 2.03  \\
OSL+RE& 1.96 & 1.99 & \textbf{49.75} & \textbf{47.39} & 2.36  \\
RSL+RE& \textbf{0.64} & \textbf{0.74} & 46.04 & 44.82 & \textbf{1.22}   \\
            \end{tabular}}
\end{table}
We first study the effect of square loss.
In addition to the rescaled square loss (RSL) we mentioned, we also compare the results when using original square loss\footnote{The original square loss is to calculate the Euclidean distance between softmax features and one-hot label vectors. Its formulation is defined as $L(\vx,\vy)=\|g(\vf(\vx))-\vy_{hot}\|$, where $g(\cdot)$ is the softmax operation.} (OSL).
We conduct experiments under the baseline training protocol. 

First, we simply replace the cross-entropy loss (CEL) with OSL. 
Fig. \ref{fig1} (a) and (b) show test accuracy curves and adversarial accuracy curves.
We observe that AT trained with CEL has a serious robust overfitting problem, where the test adversarial accuracy begins to drop after the first learning rate drops.
But training a robust model under the supervision of OSL will reduce the risk of robust overfitting, compared to that trained with CEL.

We further consider the random erasing (RE) technique.
We find that, \textit{by switching to original square loss, the learning curves are very close to the robust model trained with the combination of CEL and random erasing.} 
Furthermore, as the red curves show, the combination of OSL and random erasing can achieve better results.

We also study the effect of RSL under this baseline training protocol.
However, the model cannot be well-trained under this default.
We argue that the major problem is the larger beginning learning rate (\emph{i.e.,} $0.1$ as usual), which leads to gradient explosion.
To make the training available, we modify the beginning learning rate to $0.001$ and divide it by $10$ at the $100$-th epoch and $150$-th epoch. 
We draw the corresponding results in Fig. \ref{fig1}.
We find that RSL helps achieve a smaller degradation in both robust and natural accuracy, indicating that \textit{RSL is benefiting to mitigate robust overfiting}.

Table \ref{tab2} reports the difference between the best and final checkpoints. These results can quantitively analyze the robust overfitting.
According to the results, we find that a robust model trained with square loss (or +RE) performs better results in terms of adversarial robustness, and the gap between the highest and final checkpoints is smaller.
Note that, the model trained with RSL cannot achieve better results than that trained with CEL, but the model trained with RSL can also mitigate robust overfitting.

In sum, AT under square loss consistently mitigates robust overfitting to some extent. 
However, when compared to the previous work \cite{dong2022exploring,Pang2022RobustnessAA}, Fig. \ref{fig1}  and Table \ref{tab2} tell us the fact that: 1) using square loss does not lead to significant performance improvement; 2) the robust overfitting problem is not well addressed.

\subsection{Eﬀectiveness of Learning Rate Scheduler}
\begin{figure*}[!t]
       \centering
       \small
       \setlength\tabcolsep{-1pt}
       \begin{tabular}{ccc}
             (a) \makecell[c]{\includegraphics[width=0.32\linewidth]{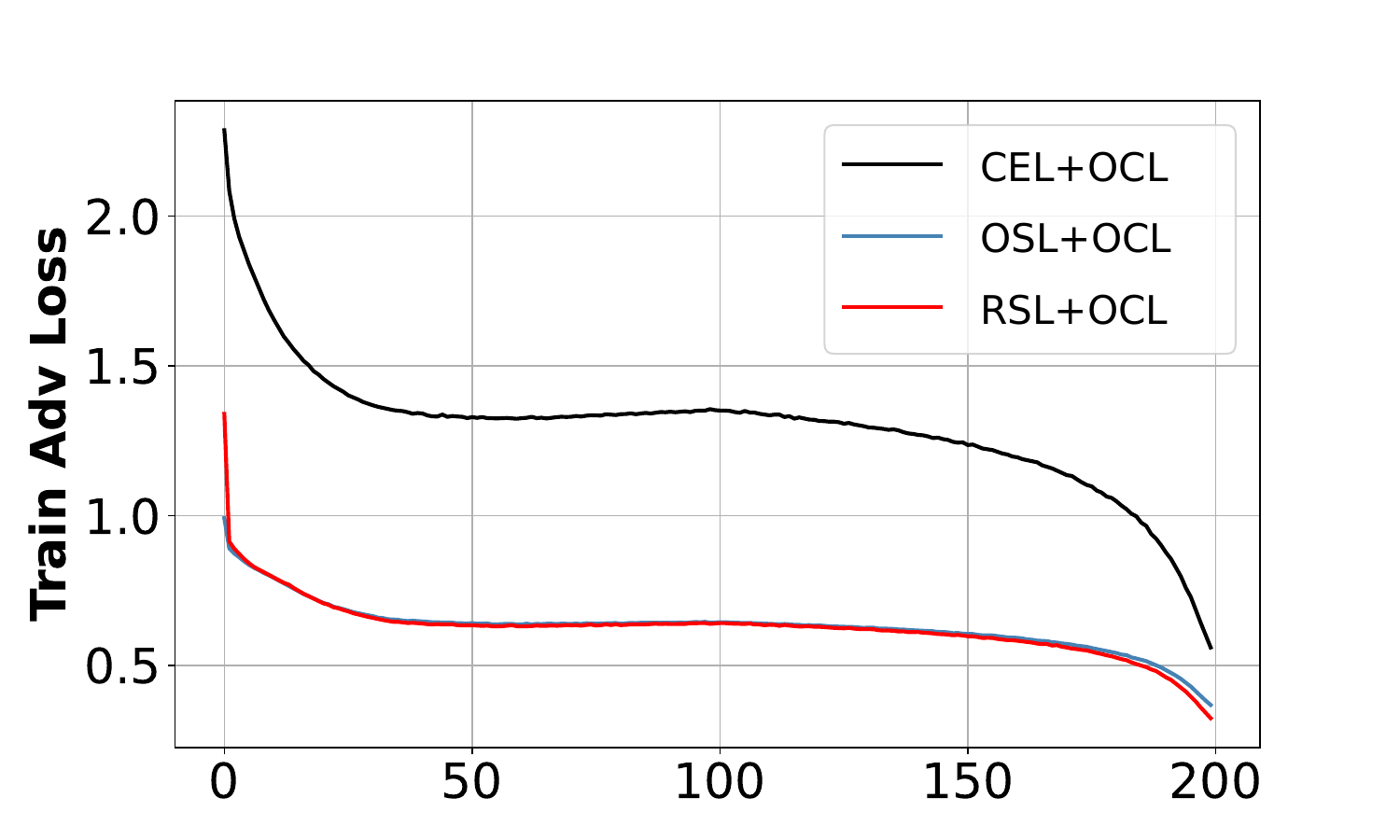}} &  
             (b) \makecell[c]{\includegraphics[width=0.32\linewidth]{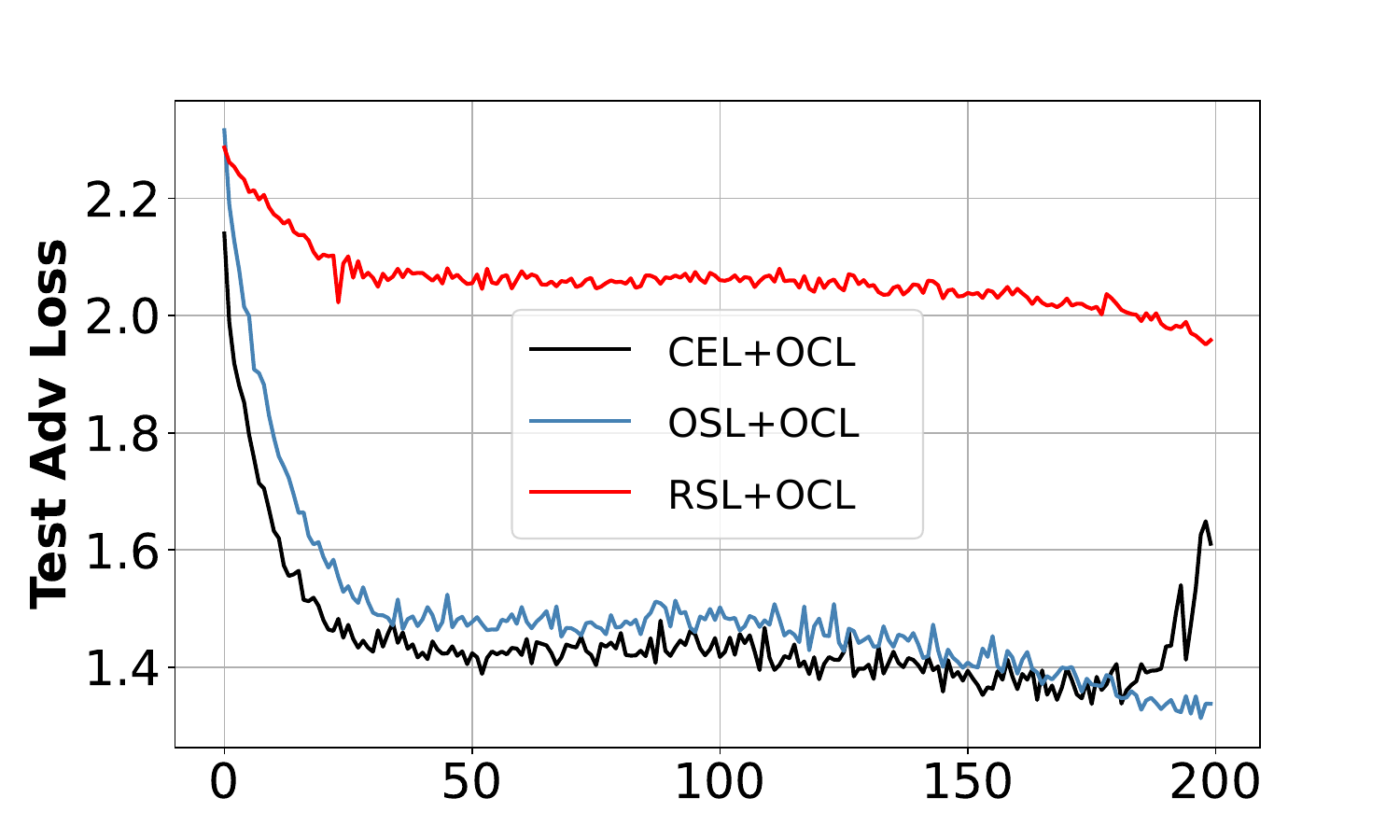}} &  
             (c) \makecell[c]{\includegraphics[width=0.32\linewidth]{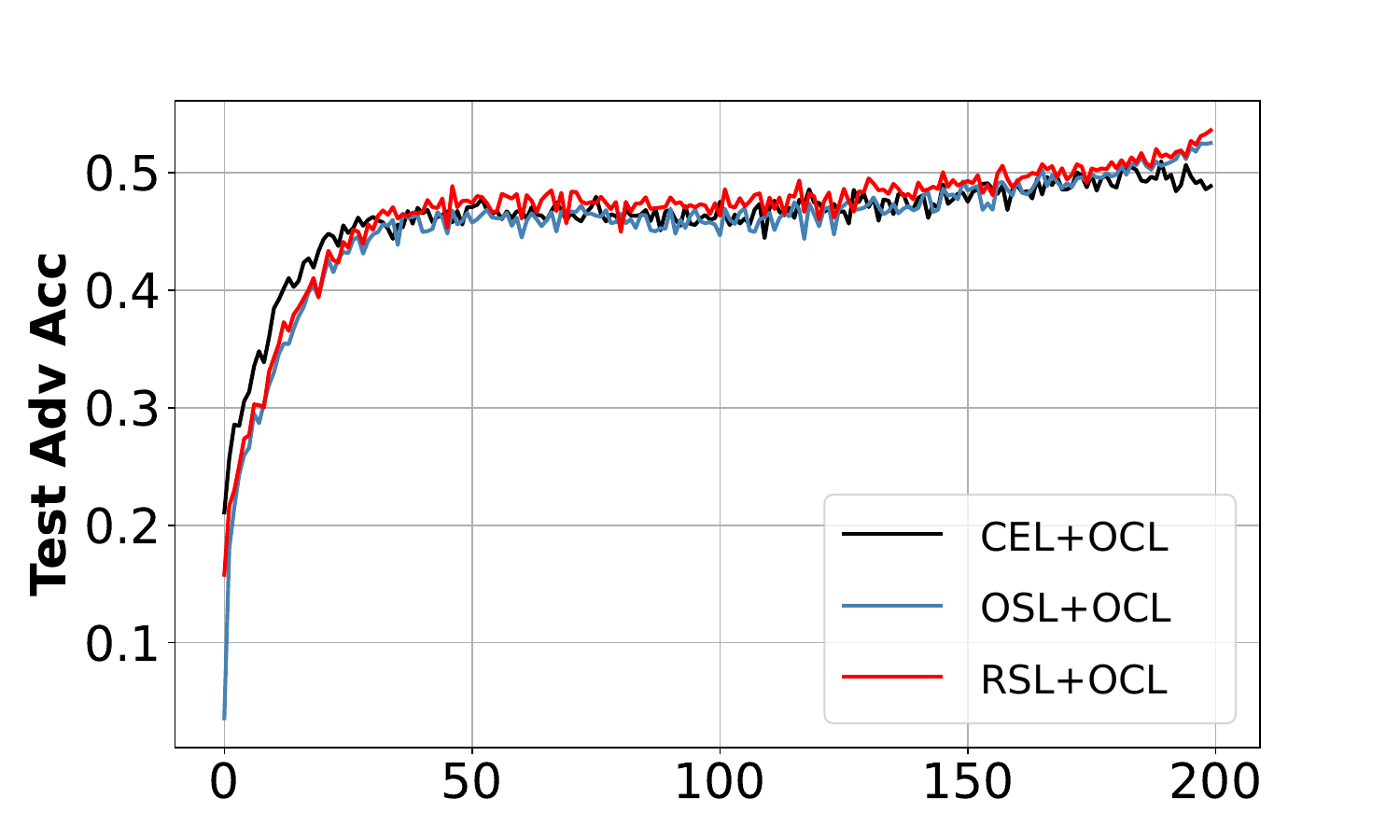} } \\
       \end{tabular} \vspace{-1.5em}
\caption{Analysis of learning rate schedulers. The results are based on a robust ResNet-18 model trained on CIFAR-10. These figures display the result curves for different loss functions under the same one cyclic learning rate scheduler.\label{fig2}} 
\end{figure*}

In previous experiments, we found that AT with RSL under baseline protocol cannot achieve good results. 
We argue that this is mainly due to the fact that we are using the wrong learning rate scheduler. 
Therefore, modifying the learning rate scheduler is crucial for achieving a robust model.
To validate this, we study the impact of learning rates through controlled experiments. 
The results are in Fig. \ref{fig2} and Table \ref{tab1}.

\begin{table}
       \centering
       \small
       \setlength\tabcolsep{2pt}
        \caption{\small Test accuracy ($\%$) on CIFAR-10. Experiments are conducted at the final epoch.\label{tab1}}
       \scalebox{1.0}[1.0]{
       \begin{tabular}[b]{lcccc}
              & Natural  & FGSM & PGD-10   & AA  \\ \midrule[1pt]
       CEL    & 82.52 & 60.44 & 48.74 & 45.19   \\
       OSL    & 84.21 & 62.30 & 52.55 & 48.04 \\
       RSL    & 84.89 & 63.41 & 53.68 & 48.59  \\
       \end{tabular}}
\end{table}
% Fig. \ref{fig2} shows results using one cyclic learning rate (OCL) scheduler to train the models on CIFAR-10.
% First, subfigure (a) shows that using the OCL schedule leads to smooth training loss curves.
% Fig. \ref{fig2} shows results using one cyclic learning rate (OCL) scheduler to train the models on CIFAR-10. 
As compared to Fig. \ref{fig1} (left), using the OCL schedule can smooth training loss curves, because the learning rate change is smooth.
In Fig. \ref{fig2} (middle), for the model trained with cross-entropy loss, the test adversarial loss continues to decrease with the training loss, however, after a certain point, the testing loss increases again, and the robust overfitting exists again.
These echo that in \cite{stutz2021relating,rice2020overfitting}, altering the learning rate scheduler cannot mitigate robust overfitting. 

On the other hand, when the model is trained with square loss (OSL or REL), the test loss continues to substantially decrease with the training loss, indicating a reduction in the robust overfitting problem.
This suggests that both one cyclic learning rate and square loss are required for training a robust model to converge well.

\begin{figure}
\centering
       \small
       \setlength\tabcolsep{-3pt}
       \begin{tabular}{cc}
       \makecell[l]{\includegraphics[width=0.53\linewidth]{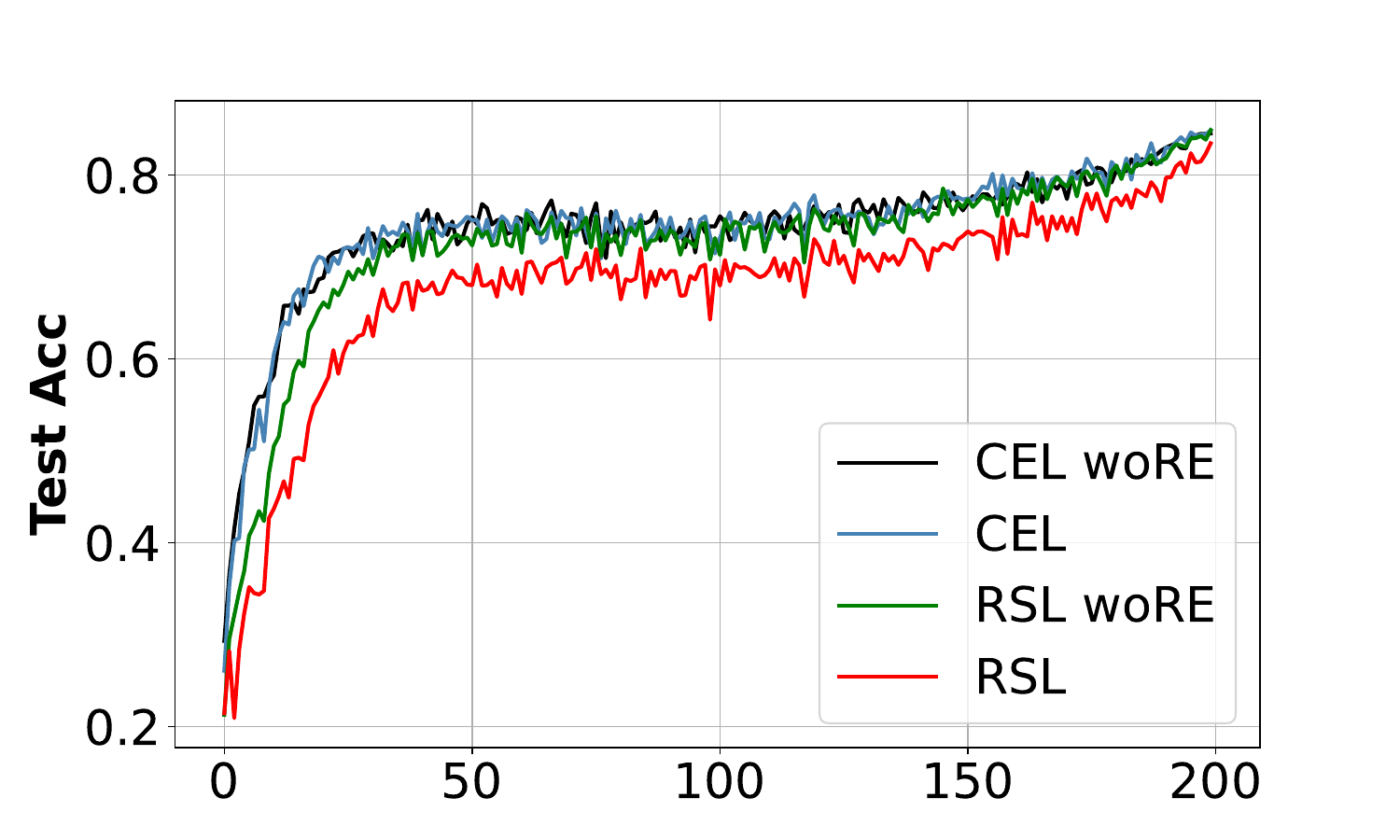} } &
       \makecell[l]{\includegraphics[width=0.53\linewidth]{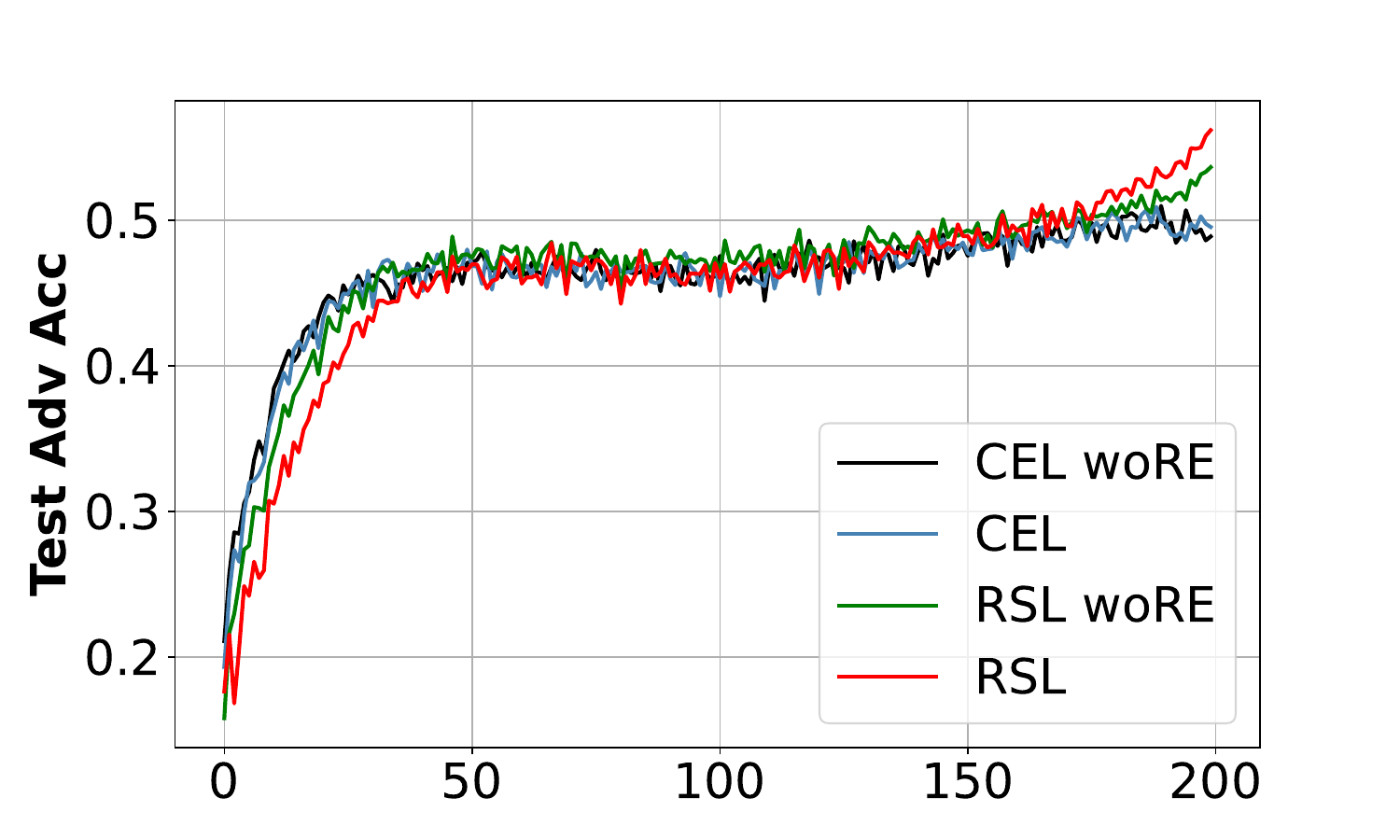} }\\
       \end{tabular}
\caption{Analysis of random erasing. Figures (a) and (b) display the effect of the default data augmentation with/without using random erasing under the baseline training protocol but with one cyclic learning rate. \label{fig3}}%\vspace{-1.5em}
\end{figure}
Fig. \ref{fig2} (right) and Table \ref{tab1} are results evaluated on CIFAR-10 test data.
The performance of the robust model trained with CEL and OCL is comparable to that of the model trained with CEL and MultiStepLR. However, employing the early-stopping technique under the baseline protocol still yields better results (\eg, 47.88\%) compared to transitioning to OCL (as reported in Table \ref{tab1}, 45.19\%).

Then, we simply replace CEL with OSL, use square loss as a training objective, and evaluate the performance on the final checkpoint. 
The results in Table \ref{tab1} show that the model achieves comparable results to those reported using the early-stopping technique under baseline training protocol.

Finally, comparing models trained with RSL and OSL, we observe that RSL contributes to further performance improvement, whether in natural accuracy or adversarial accuracy.
Meanwhile, the results in Fig. \ref{fig2} indicate that when using OCL, RSL also helps alleviate robust overfitting problems.

Overall, \textit{modifying the baseline training protocol to include rescaled square loss and one cyclic learning rate is all we needed}. This modification can achieve good results, while also neglecting the disreputable overfitting problem.

\begin{figure*}[!t]
       \centering
       \small
       \setlength\tabcolsep{-1pt}
       \begin{tabular}{ccc}
       (a) \makecell[l]{\includegraphics[width=0.32\linewidth]{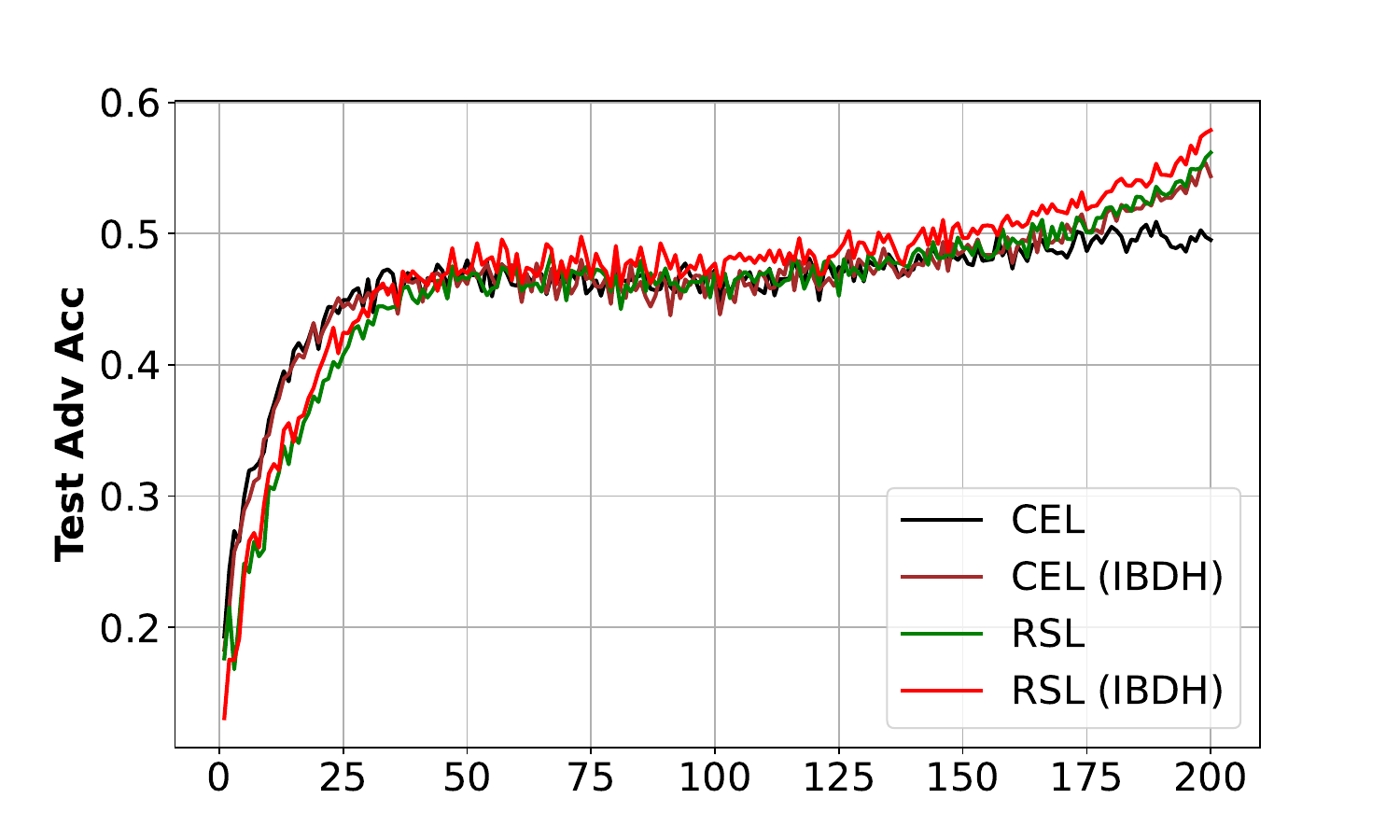} } &
       (b) \makecell[l]{\includegraphics[width=0.32\linewidth]{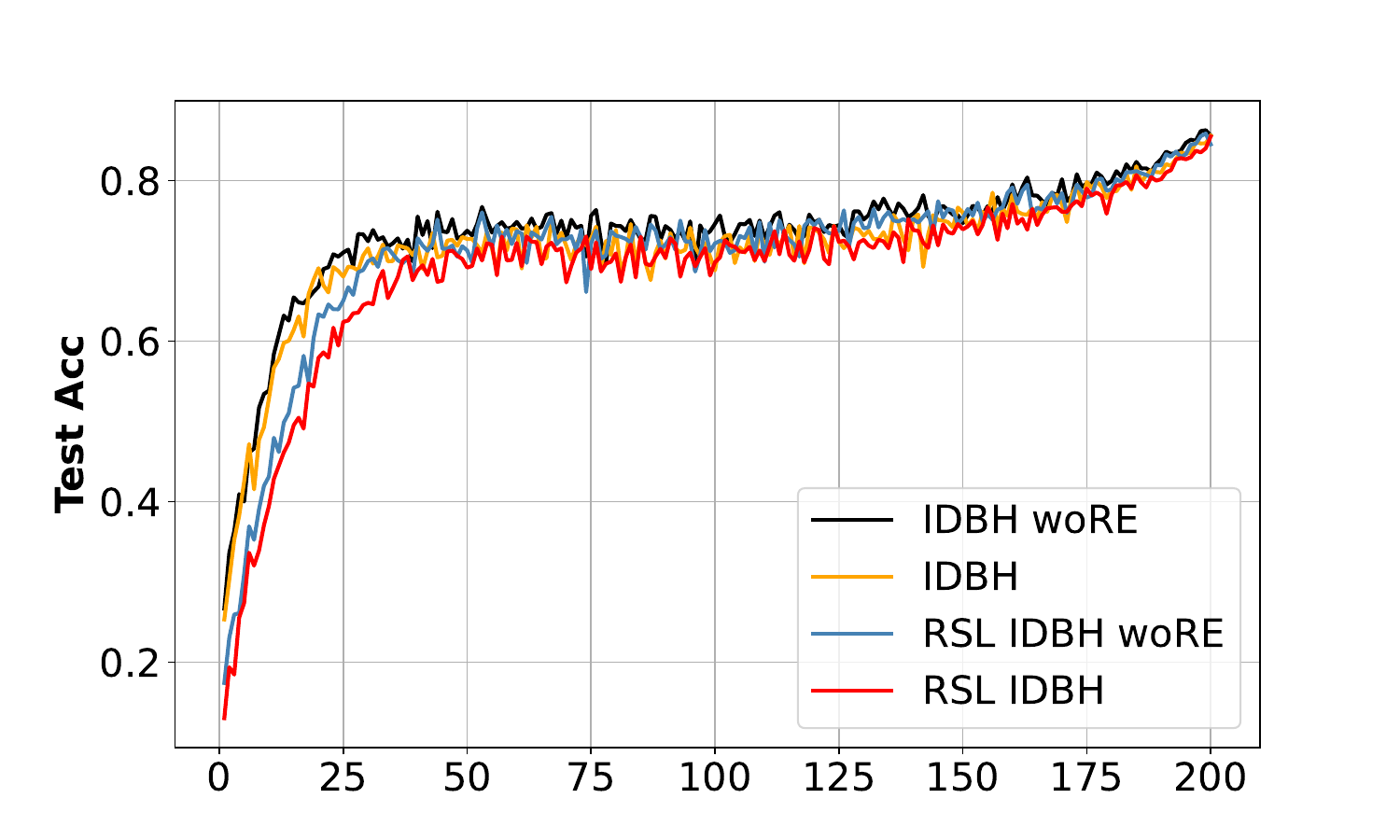} } &
       (c)\makecell[l]{\includegraphics[width=0.32\linewidth]{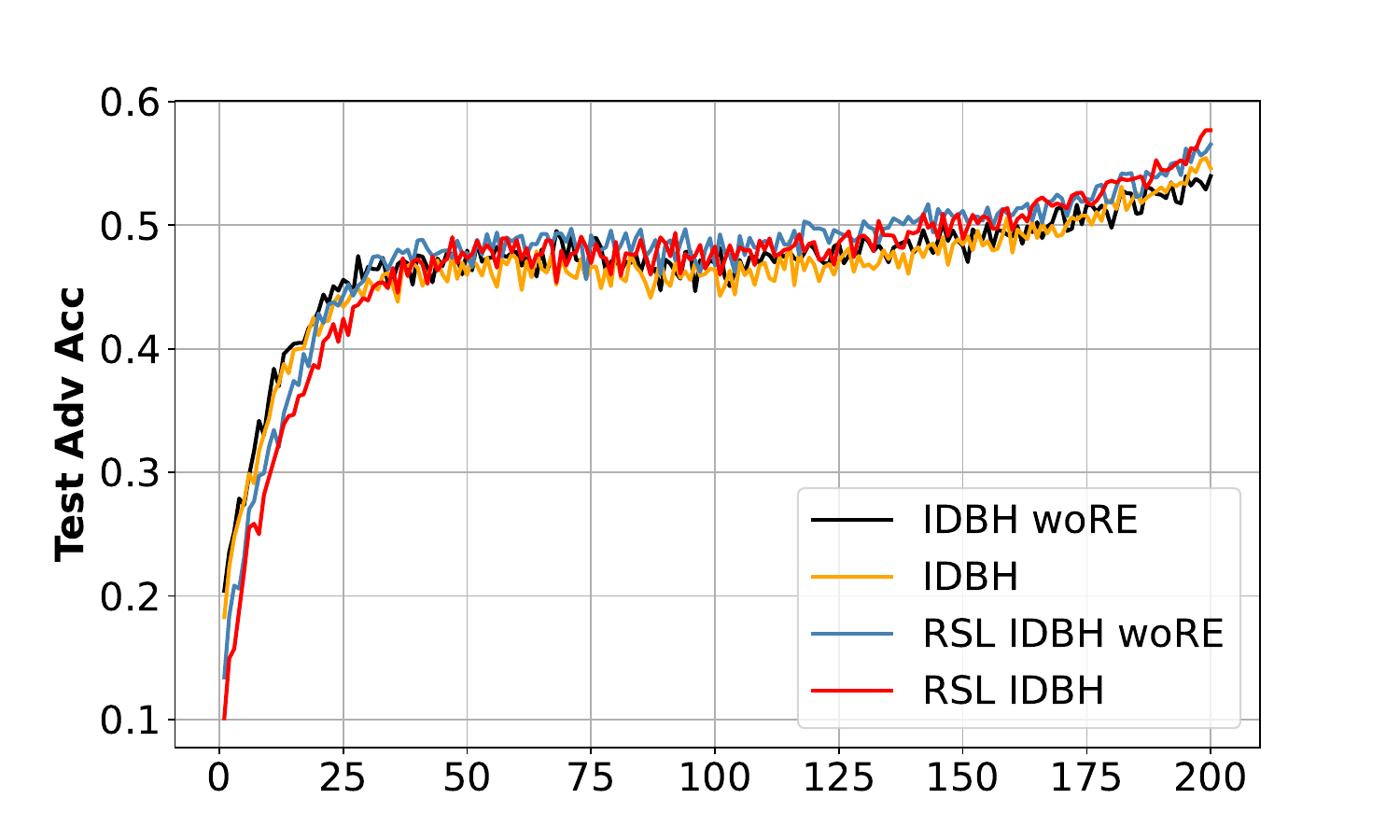} } \\
       \end{tabular} \vspace{-1.5em}
\caption{Analysis of random erasing. All results are evaluated on a robust ResNet-18 trained on CIFAR-10. Subfigure (a) studies the effect of the IDBH. Subfigure (b) and (c) show the test accuracy curves when we study the influence of random erasing modules in the IDBH scheme. 
 \label{fig4}}%\vspace{-1.5em}
\end{figure*}

\subsection{Effectiveness of Data Augmentation}

\begin{table*}[h]
       \centering
       \footnotesize
       \setlength\tabcolsep{8pt}
       \caption{Analysis of robust overfitting. Test accuracy (\%)  of several models trained on CIFAR-10. The backbone we used here is ResNet-18. The ``Natural'' is the natural accuracy. The ``Best'' reports the results of the checkpoint and who achieves the best performance during the training. The ``Final'' is the results of the last training epoch. The ``Diff'' shows the difference between the best and final checkpoints.  \label{RO}}
       \begin{tabular}{rcccccccccc}
              \multirow{2}{*}{Method} & \multirow{2}{*}{Protocol} &  \multicolumn{3}{c}{Natural} & \multicolumn{3}{c}{PGD-10} & \multicolumn{3}{c}{AutoAttack} \\
       \cmidrule(l){3-5} \cmidrule(l){6-8} \cmidrule(l){9-11} \\
                            &   & Best    & Final    & Diff   & Best    & Final   & Diff   & Best     & Final     & Diff    \\\midrule[1pt]
        PGD-AT  & \multirow{6}{*}{Default} & 81.48   & 84.00    & 2.52   & 52.06   & 44.92   & 7.14   & 47.88    & 41.07     & 6.81    \\
       % PGD-AT-LS               & 82.68   & 85.16    & 2.48   & 53.70   & 48.90   & 4.80   & 49.02    & 44.39     & 4.63    \\
       SAT  &  & 82.81   & 81.86    & 0.95   & 53.81   & 53.31   & 0.50   & 50.21    & 49.73     & 0.48    \\
        KD-SWA    &    & 84.84   & 85.26    & 0.42   & 54.89   & 53.80   & 1.09   & 50.42    & 49.83     & 0.59    \\
       Co-teaching    &         & 81.94   & 82.22    & 0.28   & 51.27   & 50.52   & 0.75   & 49.60    & 48.49     & 1.11    \\
       TE    &       & 82.35   & 82.79    & 0.44   & 55.79   & 54.83   & 0.96   & 50.59    & 49.62     & 0.97    \\
       SRC &   & 80.30   & 80.70    & 0.40   & 57.55   & 57.90   & 0.35   & 50.38    & 50.35     & 0.03    \\ \midrule[0.5pt]
        TRADES &  & 79.17   & 80.63    & 1.46   & 56.80   & 56.43   & 0.46   & 49.07    & 49.54     & 0.47    \\ 
       SCORES     &              & 80.84   & 80.84    & 0.00   & 56.06   & 56.06   & 0.00   & 50.14    & 50.14     & 0.00    \\ 
       IDBH         &             & 84.64   & 85.59    & 0.95   & 55.39   & 54.61   & 0.78   & 50.66    & 50.00     & 0.66    \\ \rowcolor{mygray!50}
       SimpleAT    &  & 83.58   & 83.58    & 0.00   & 56.22   & 56.22   & 0.00   & 50.61    & 50.61     & 0.00    \\ \rowcolor{mygray!50}
       SimpleAT (IDBH)  &   & 84.84   & 84.84    & 0.00   & 57.70   & 57.70   & 0.00   & 51.54    & 51.54     & 0.00    \\  \rowcolor{mygray!50}
       SimpleAT (IDBH+WA) &  \multirow{-6.5}{*}{Ours}   & 85.33   & 85.33    & 0.00   & 58.38   &  58.38   &  0.00   &  52.30    &  52.30     &  0.00    \\ 
       \end{tabular}
\end{table*}

In this subsection, we study the influence of data augmentation.
We compare two different data augmentation methods, both of which contain random erasing.
The first default data augmentation utilizes random crop, random flip, and random erasing. 
The second one is a recent state-of-the-art data augmentation method, called IDBH, which consists of CropShift, ColorShape, and random erasing. 
We follow the default setting recommended by \cite{li2023data}.

In Fig. \ref{fig3}, we draw the result curves under the default data augmentation.
We find that the best performance is achieved at the final phase of training for both natural and adversary accuracy.
However, the robust model trained with CEL is unstable at this phase, with even minor performance degradation.
In other words, the random erasing technique does not provide significant assistance for models trained using cross-entropy loss.
In Fig. \ref{fig3}, we also find that using (rescaled) square loss with random erasing can achieve better robust performance while maintaining natural accuracy on par with the model trained with CEL. 

Fig. \ref{fig4} shows the results, when random erasing is used or not used in IDBH.
We observe that all methods perform best during the final phase of training, where robust overfitting is mitigated well.
Fig. \ref{fig4} (a) indicates that IDBH is an effective data augmentation for robustness enhancement. 
Then, under the setting of IDBH, the difference in natural accuracy between different methods is not very large, whether random erasing is used or not, as shown in Fig. \ref{fig4} (b). 

Fig. \ref{fig4} (c) shows that using random erasing improves robust accuracy for both CEL and RSL.
But, when using CEL and random erasing, we find that the robust accuracy has a little degradation at the final checkpoints, although it is better than that trained without random erasing.
This phenomenon does not exist when we use RSL as the supervisor.  
That is, using both RSL and random erasing can not only achieve good results but also reach a good trade-off between accuracy and robustness.
Previous works show that robustness improvement will bring accuracy drop, but our observation shows that ours can improve robustness without accuracy degradation.

Overall, combining the random erasing and OCL also mitigates robust overfitting under CEL for standard AT.
Therefore, \textit{modifying training protocols to include random erasing and one cyclic learning rate is all that is needed.} We mainly use IDBH as our data augmentation method, which helps us to achieve good results.

\section{Comparisons with Previous Works\label{sec4}}

In the previous section, we conducted detailed ablation studies using controlled experiments. The results show that our proposed training protocol (see Table \ref{SVP}) is competitive when compared to previous default protocols.
In this section, we will compare the results with those of many recent state-of-the-art methods under the PGD-based AT paradigm. 
It's worth noting that our robust model is trained only under the proposed training protocol without adding other regularizations or modifying the architectures.

We conduct an analysis of the problem of robust overfitting for various methods, including SAT \cite{huang2020self}, KD-SWA \cite{chen2021robust}, TE \cite{dong2022exploring}, and SRC \cite{Liu2023MitigatingRO}, which aim to mitigate robust overfitting. 
In addition, we compare various regularization and data augmentation methods, such as TRADES \cite{zhang2019theoretically}, SCORES \cite{Pang2022RobustnessAA}, and IDBH \cite{li2023data}, trained under our proposed training protocol.
All the training protocols can be found in Table \ref{SVP}.

Table \ref{RO} presents the results on CIFAR-10, showing that all methods, except for the \textit{de-facto} PGD-AT, perform well in reducing the risk of robust overfitting. 
Under our training protocol, we observe that classical TRADES can achieve competitive results in the final stage of training when compared to more recent techniques such as TE and SRC.
Then, by simply replacing the KL divergence with the Euclidean distance, the SCORES can further improve the performance, and the robust overfitting problem is well mitigated.
Although both TRADES and SCORES exhibit good robustness, this is achieved by sacrificing a certain degree of natural accuracy, namely facing a trade-off dilemma.

Interestingly, SimpleAT achieves the best performance at final checkpoints and maintains a good balance between natural and adversarial accuracy, surpassing all compared methods.   
Moreover, the utilization of IDBH and weight averaging (WA) techniques leads to a significant enhancement in overall performance. 
Our approach surpasses most cutting-edge methods, as evidenced by the last three rows in Table \ref{RO}. 
For example, the adversarial accuracy against AutoAttack reaches 52.30\%, which is a notable result when the backbone is ResNet-18.

\begin{table*}[t]
       \centering
       \footnotesize
       \setlength\tabcolsep{4pt}
       \caption{Performance of several models trained on CIFAR-10. The backbone is WideResNet. \label{WN}}
       \begin{tabular}{lcccccccccgg}
       Method     & TRADES & ES & SAT & FAT & LBGAT & AT-HE & Tricks & LTD  & IDBH & SimpleAT & SimpleAT (+WA) \\\midrule[1pt]
       Natural    &  84.92      &  85.35  &   83.48  &  84.52   &  88.70 & 85.14 & 86.43 & 85.02 & 89.93 & 90.04 & 89.82          \\
       AutoAttack &   53.08     &  53.42  & 53.34   & 53.51    &  53.57     &   53.74    &      54.39  &   54.45  & 54.10  & 54.55  &  55.95               
       \end{tabular}
       
\end{table*}
We evaluate the performance when using a larger network, \ie,~WideResNet, in which we compare our SimpleAT with various recent robust models, including SAT \cite{huang2020self}, FAT \cite{zhang2020attacks}, LBGAT \cite{cui2021learnable}, AT-HE \cite{pang2020boosting}, Tricks \cite{pang2021bag}, LTD \cite{chen2021ltd}, IDBH \cite{li2023data}. 
Similar to \cite{rice2020overfitting}, we also use an early-stopping strategy (ES) to avoid robust overﬁtting, which can also achieve good results.
The results are reported in Table \ref{WN}.
We find that our SimpleAT consistently outperforms all the compared methods in terms of both natural accuracy and robust performance. Echoing the results on ResNet-18, the robustness can be further improved by using weight averaging techniques. Although there is a slight degradation in the natural accuracy, it is competitive since the 89.82\% score is better than most of the compared methods and only lower than IDBH.

\begin{table}[!t]
       \centering
       \setlength\tabcolsep{3pt}
       \caption{Test accuracy (\%) on CIFAR-10 when using extra dataset during adversarial training.\label{extra}}
       \begin{tabular}{llccccc}
              Arch & Method & Generated & Batch & Epoch & Clean & AutoAttack \\\midrule[1pt]
              \multirow{4}{*}{WRN-28-10}  & \cite{rebuffi2021data} & 1M & 1024 & 800 & 87.33 & 60.73 \\
              & \cite{Pang2022RobustnessAA} & 1M & 512 & 400 & 88.10 & 61.51 \\
              & \cite{Wang2023BetterDM} &1M & 512 & 400 & 91.99 & 61.46  \\ 
         \rowcolor{mygray!50}    \cellcolor{white}         & SimpleAT & 1M & 512 & 400 & 92.30 & 61.95 \\\midrule[0.5pt]
 \multirow{1}{*}{WRN-76-16} & SimpleAT &  5M & 512 & 400 & 93.08 & 65.62\\
       \end{tabular}\\
\end{table}

Recently, researchers have used extra datasets to enhance model robustness, resulting in top-tier performance on the RobustBench leaderboard \cite{rebuffi2021data,Pang2022RobustnessAA,Wang2023BetterDM}.
We follow the experimental setup in \cite{Wang2023BetterDM}, which employs a diffusion model to generate 1 million virtual samples.
 % sampled from a larger pool of 5 million virtual images.
We default to using WideResNet-28-10 as the backbone.
See Table \ref{extra}. Our SimpleAT performs better with additional data for training, achieving better robustness against AutoAttack and significantly improving natural accuracy. 
% For instance, SimpleAT shows a clean data accuracy gain of about 1.43\% when compared to \cite{Wang2023BetterDM}.
Finally, we evaluate performance using a larger backbone, \ie, WideResNet-76-16.
SimpleAT can reach 65.53\% adversarial accuracy against AutoAttack and 93.08\% clean accuracy.
This means that SimpleAT can achieve the top-rank performance on the RobustBench leaderboard.
However, due to our limited computational resources, we are not able to further scale up the data amount to enhance robustness, which may take months to train one model. We believe that our SimpleAT can achieve better results when using the larger extra dataset (like with 50M images).

\section{Experiments on FGSM-based AT methods\label{sec5}}

\begin{figure}[!t]
       \centering
       \footnotesize
       \setlength\tabcolsep{-3pt}
       \begin{tabular}{ccc}
              \makecell[c]{\includegraphics[width=0.54\linewidth]{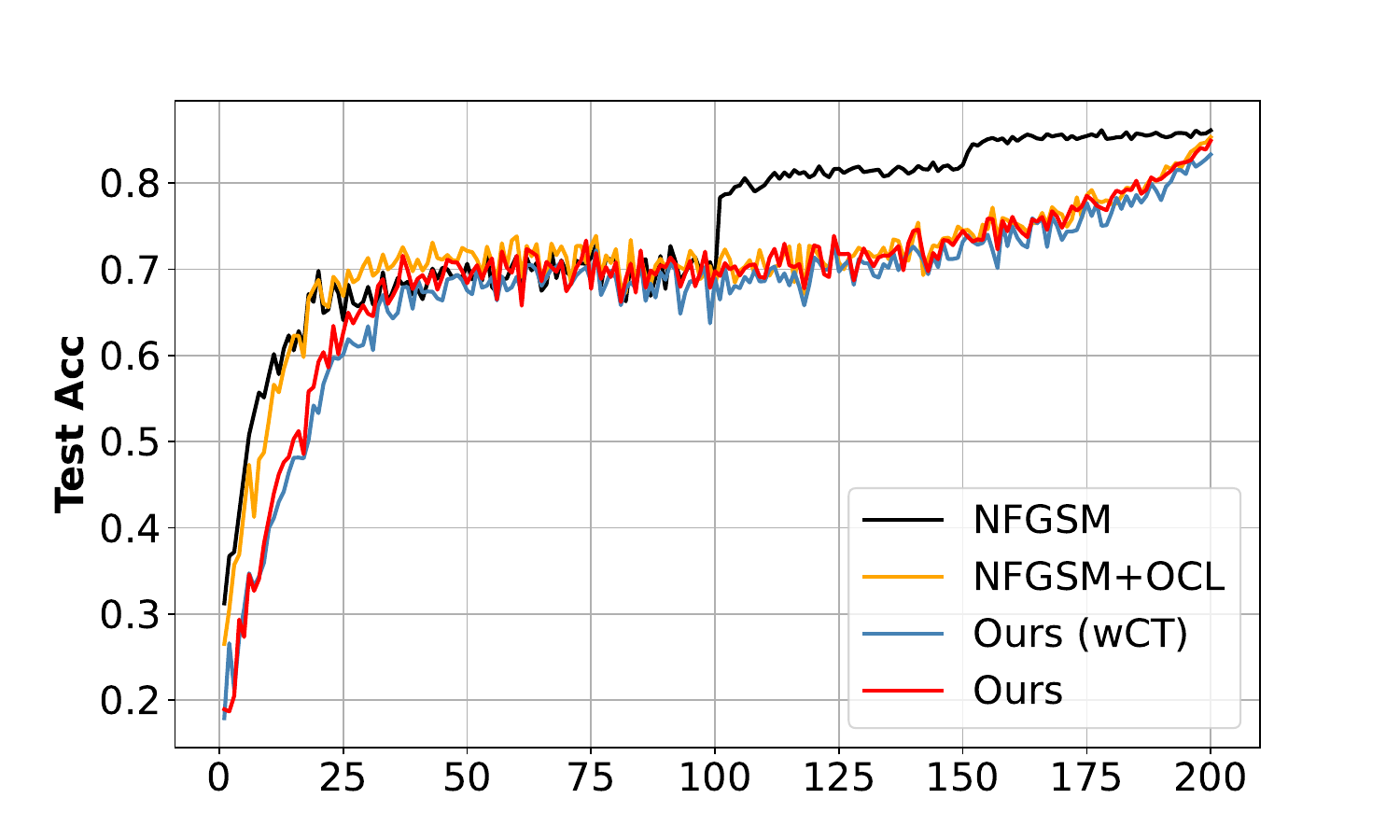}} &  
              \makecell[c]{\includegraphics[width=0.54\linewidth]{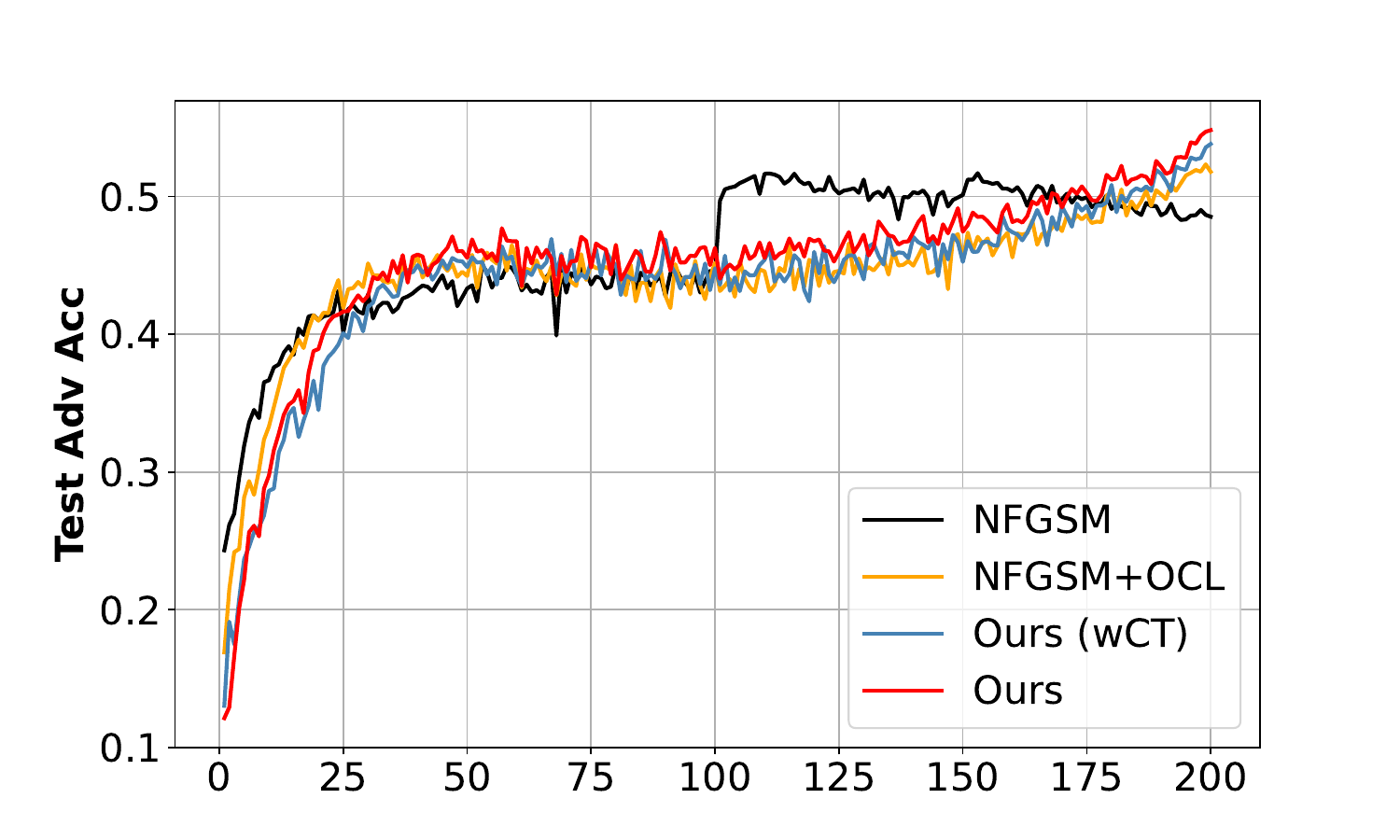}} & 
%              \begin{tabular}{rccccc}
%  & NFGSM & PGD  \\\midrule[1pt]
% Natural & 60.74 & 60.89 \\
% FGSM & 39.46  & 40.51  \\
% PGD-10 & 32.86  & 34.60  \\
% AA &  26.57 & 27.89  \\ \hline
% Time (h) &  2.03 & 10.34  \\
% \multicolumn{3}{c} {\footnotesize(c) Performance on CIFAR-100.}\\
%               \end{tabular}\\
       \end{tabular}\\
       \caption{Analysis of FGSM-based AT methods. All models are trained on CIFAR-10. The backbone is ResNet-18. Figure (a) shows the clean accuracy curves and figure (b) draws the adversarial accuracy curves. \label{FA1}}
\end{figure}

This section mainly evaluates the performance under a more effective FGSM-based AT paradigm.
We compare three baseline methods, such as AT-GA \cite{andriushchenko2020understanding}, FAST-BAT \cite{zhang2021revisiting}, and NFGSM \cite{jorge2022make}.
We report our results in Fig. \ref{FA1} and Fig. \ref{FA2}, which are evaluated on CIFAR-10/100 and Tiny-ImageNet.

In Fig. \ref{FA1} (a) and (b), we study whether SimpleAT suffers from catastrophic overfitting.
Echoing the results shown in \cite{jorge2022make}, NFGSM makes the adversarial training stable, but it still faces robust overfitting under such a longer training scheme.
Therefore, we changed the learning rate schedule to one cyclic learning rate (OCL). 
The risk of robust overfitting can be reduced, similar to PGD-based AT, as Fig. \ref{FA1} (b) shows.
After comparing the yellow and blue curves in Fig. \ref{FA1}, we find that RSL improves the robustness but slightly degrades natural accuracy. 
Based on the previous experience, we further use IDBH for the data augmentation, resulting in improved natural accuracy that is comparable to the CEL-trained model.
Moreover, we also find that the robust accuracy against PGD-10 attack is 54.77\%, which is comparable to that model trained with PGD-based AT. 

To further verify this, we conduct similar experiments on CIFAR-100 and report the results:
\begin{center}
\small
\setlength\tabcolsep{3pt}
\begin{tabular}{rccccc}
         & Natural & FGSM  & PGD-10 & AA    & Times (h) \\ \midrule[1pt]
NFGSM-AT & 60.74   & 39.46 & 32.86  & 26.57 & 2.03      \\ 
PGD-AT   & 60.89   & 40.51 & 34.60  & 27.89 & 10.34     \\ 
\end{tabular}
\end{center}
A similar phenomenon is observed: SimpleAT is effective and performs similar results to PGD-based AT. 
It's worth noting that PGD-based AT requires 5 times more training time.

\begin{figure}[!t]
       \centering
       \footnotesize
       \setlength\tabcolsep{-4pt}
       \begin{tabular}{ccc}
              \makecell[c]{\includegraphics[width=0.54\linewidth]{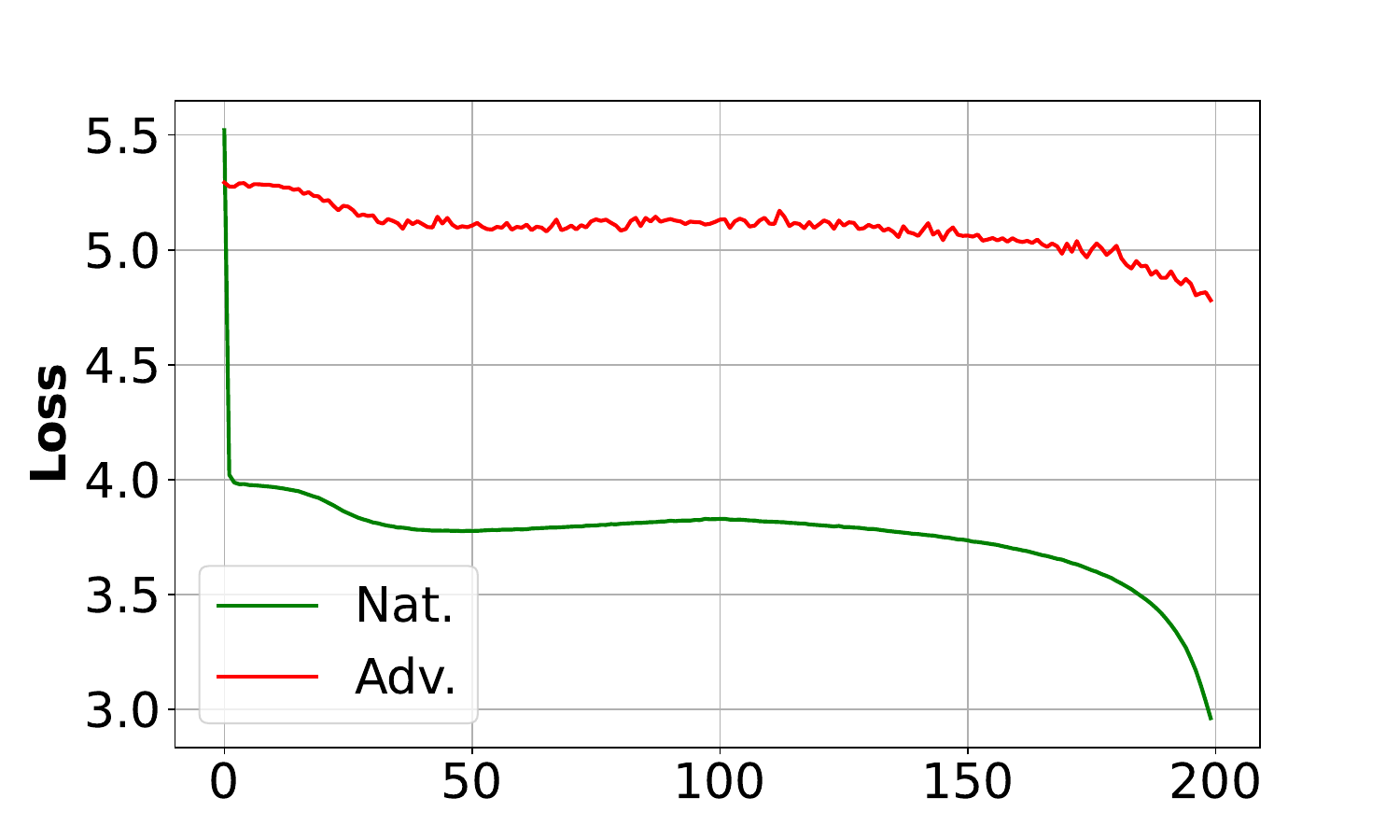}} &  
              \makecell[c]{\includegraphics[width=0.54\linewidth]{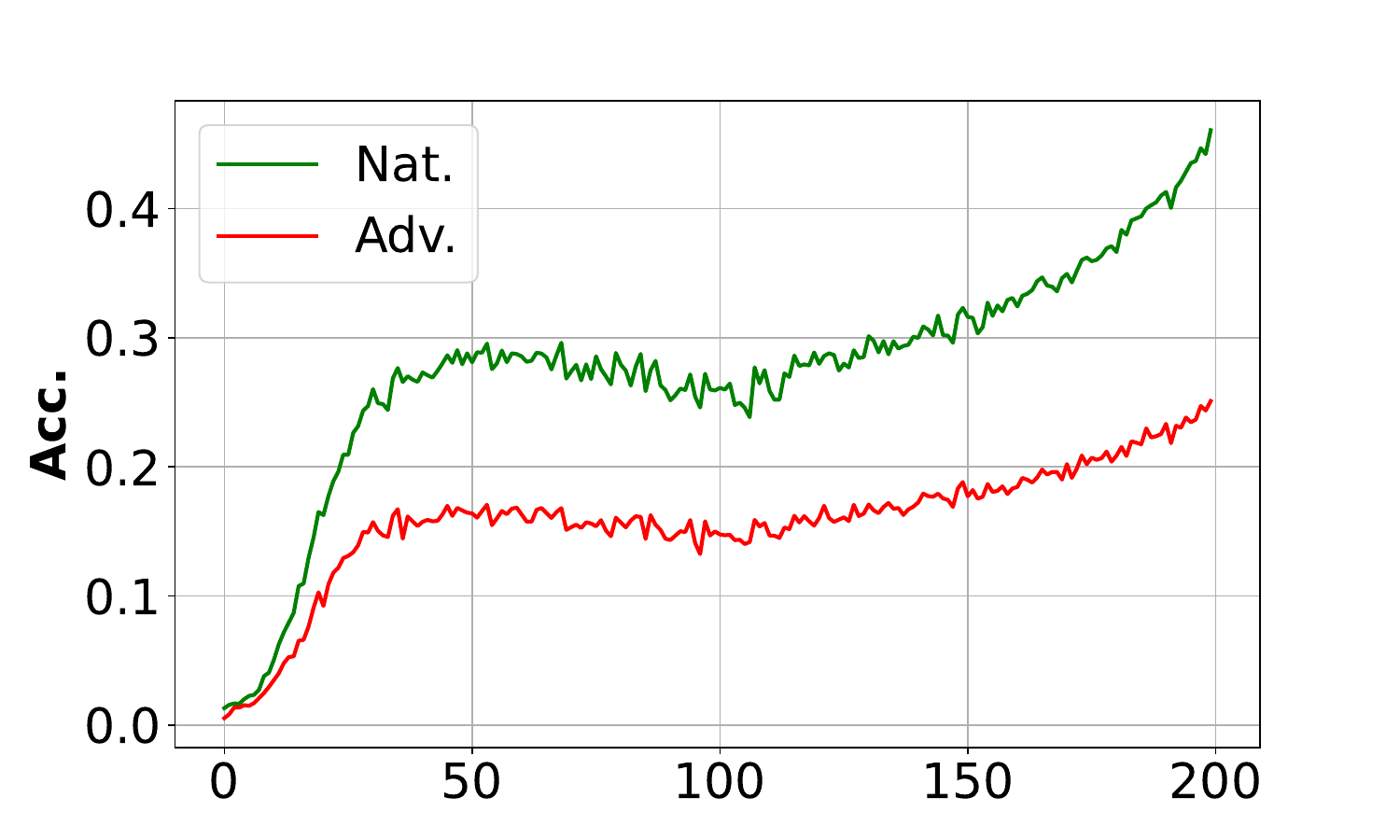}} & 
%              \begin{tabular}{lccccc}
% Method & Natural & PGD & AA  \\\midrule[1pt]
% FGSM & 41.37 & 17.05 & 12.31\\
% AT-GA & 45.52 & 20.39 & 16.25  \\
% % SRC & 46.68  & 21.36 & 16.97   \\
% FAST-BAT & 45.80  & 21.97 & 17.64  \\
% % SRC+FL & 46.75  & 22.06 & 17.01   \\
% Ours & 46.08  & 25.05 & 18.68  \\
% \multicolumn{4}{c} {\small (c) Quantitive results.}\\
%               \end{tabular}\\
       \end{tabular}\\
       \caption{Compared results of different FGSM-based methods trained on Tiny-ImageNet. The backbone is ResNet-18. Subfigure (a) shows the learning loss curves and subfigure (b) draws the test accuracy curves.\label{FA2}}
\end{figure}
Fig. \ref{FA2} shows the results on a large real-world dataset, \ie, Tiny-ImageNet, where we compare SimpleAT with three high-effect AT methods such as FGSM \cite{Wong2020Fast}, AT-GA \cite{andriushchenko2020understanding}, FAST-BAT \cite{zhang2021revisiting}, and SRC \cite{Liu2023MitigatingRO}.
Since the original IDBH does not provide recommended hyper-parameter settings for Tiny-ImageNet, we use RandomCrop, RandomHorizontalFlip, and random erasing as the data augmentation scheme on Tiny-ImageNet.
Fig. \ref{FA2} (left) displays the learning loss curves for natural and adversarial images, while Figure \ref{FA2} (right) shows the test accuracy curves during training. 
Our method ensures stability in FGSM-AT by preventing both robust and catastrophic overfitting.

We further show the quantitive results as follows:
\begin{center}
\small
\setlength\tabcolsep{3pt}
\begin{tabular}{lccccc}
Method & Natural & PGD & AA  \\\midrule[1pt]
FGSM & 41.37 & 17.05 & 12.31\\
AT-GA & 45.52 & 20.39 & 16.25  \\
SRC & 46.68  & 21.36 & 16.97   \\
FAST-BAT & 45.80  & 21.97 & 17.64  \\
% SRC+FL & 46.75  & 22.06 & 17.01   \\
Ours & 46.08  & 25.05 & 18.68  \\
\end{tabular}
\end{center} 
SimpleAT is consistently better than all compared methods.   
Specifically, compared to FAST-BAT, SimpleAT reaches gains of $0.6\%$, $14.02\%$, and $12.24\%$ on natural, PGD-50, and AutoAttack, respectively. 
In other words, SimpleAT is able to substantially improve robustness while maintaining high natural accuracy. 
These results verify our claim that SimpleAT yields a favorable trade-off balance between adversarial and natural accuracy. 
\begin{table*}[!t]
       \centering
%       \footnotesize
       \setlength\tabcolsep{5pt}
       \caption{Evaluation on CIFAR-10-C. We report the accuracy of ResNet-18 models trained on the original CIFAR-10. \label{C10C} }
       \begin{tabular}{lccccccccg}
       Method     & Standard & 100\% Gauss & 50\% Gauss & Fast PAT & AdvProp & $l_{\infty}$ adv. & $l_2$ adv. & RLAT  & SimpleAT  \\\midrule[1pt]
       Standard Acc.    &  95.1     &  92.5  &   93.2  &  93.4   &  94.7 & 93.3 & 93.6 & 93.1 & 94.0         \\
       Corruption Acc. &   74.6     &  80.5  & 85.0   & 80.6   &  82.9    &   82.7    &   83.4  &   84.1  & 87.2               
       \end{tabular}
\end{table*}

\section{Further Experiments on CIFAR-10-C\label{sec6}}

All foregoing experiments are conducted under adversarial attacks, which mainly reflect the worst-case performance for an evaluated model.
However, in the real world, there are so many other types of corruption that affect the model performance, like noise, blur, digital, and weather corruption.
To further verify the effectiveness of our method, we conduct experiments on CIFAR-10-C.
Following recent results in \cite{kireev2022effectiveness}, we use a small perturbation budget where $\epsilon=2/255$.
We calculate the average accuracy over 15 different common corruptions, which show the average-case behavior of the robust model.
We compare eight different baseline methods, including standard training, $l_2$ and $l_\infty$ adversarial training, Gaussian augmentation (with both 50\% and 100\%), AdvProp \cite{xie2020adversarial}, PAT \cite{laidlaw2020perceptual}, and RLAT \cite{kireev2022effectiveness}.
See Table \ref{C10C}.
SimpleAT also achieves the best robust performance, which has a 3.69\% gain compared to the second-best RLAT.
Although the standard accuracy is not the best, we think our result of 94.0\% is also competitive.

\section{Further Analysis \label{sec7}}

% In this section, we provide a comprehensive explanation of why SimpleAT performs well from different perspectives: adversarial bias-variance decomposition (see \ref{sec41}) and kernel methods (see \ref{sec42}). 

\subsection{Bias-variance decomposition\label{sec41}}
\begin{figure*}[!t]
       \centering
       \footnotesize
       \setlength\tabcolsep{0pt}
       \begin{tabular}{cccc}
       %        \makecell[c]{\includegraphics[width=0.25\linewidth]{./fig/fig8/ver_std.pdf}}
       %      & 
       %     \makecell[c]{\includegraphics[width=0.25\linewidth]{./fig/fig8/ver_adv.pdf}} &
            (a) \makecell[c]{\includegraphics[width=0.3\linewidth]{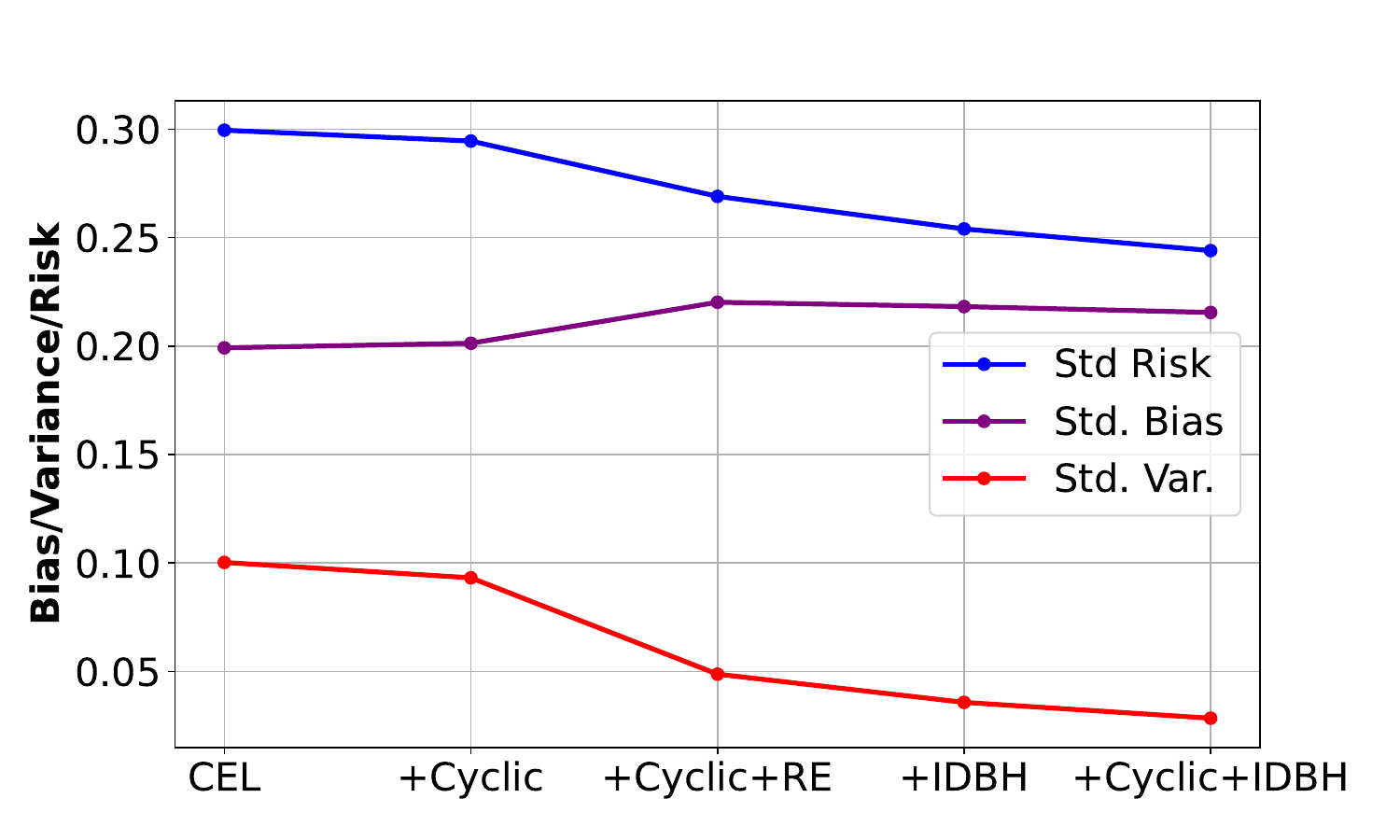}} &
            (b) \makecell[c]{\includegraphics[width=0.3\linewidth]{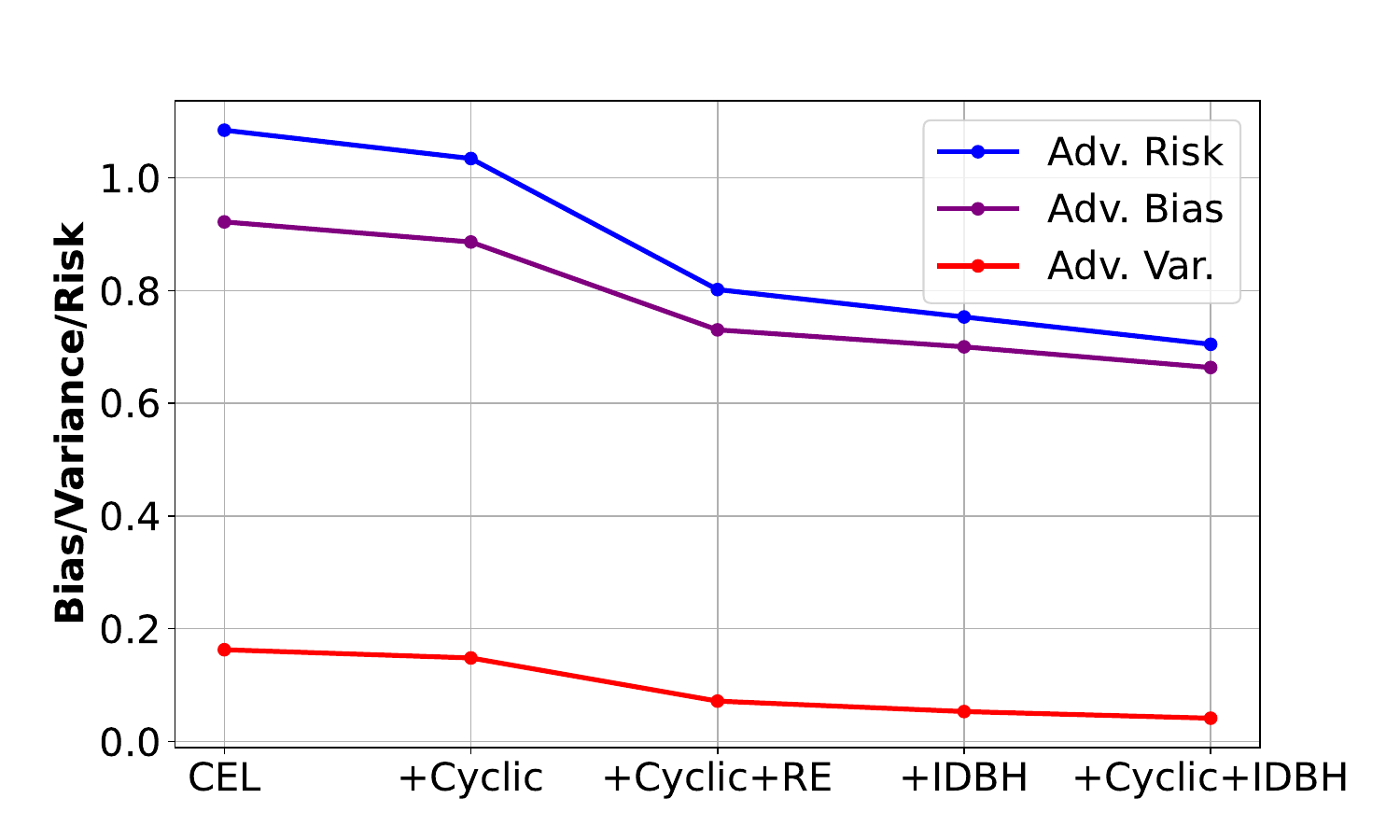}} &
            (c) \makecell[c]{\includegraphics[width=0.3\linewidth]{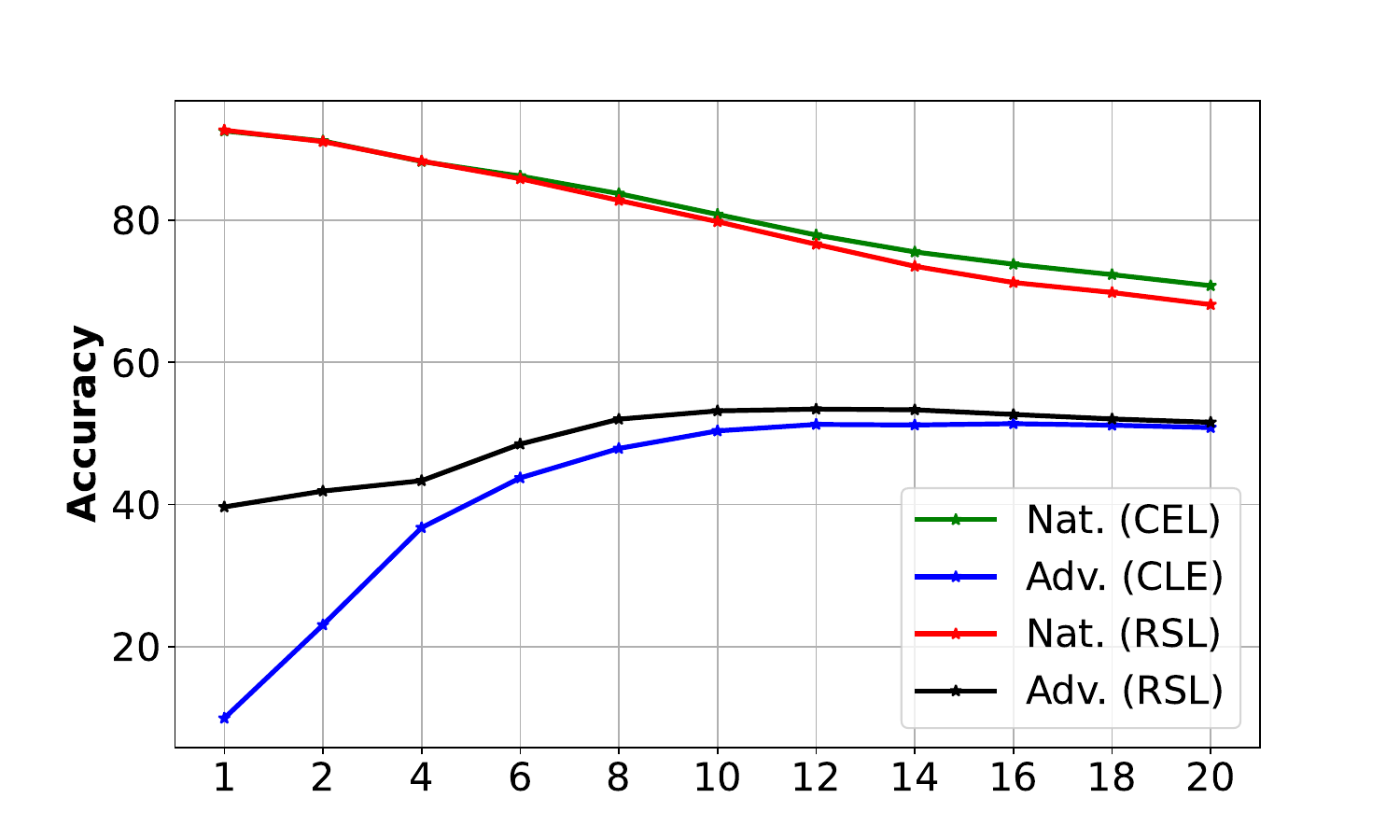} }\\
           (d)  \makecell[c]{\includegraphics[width=0.3\linewidth]{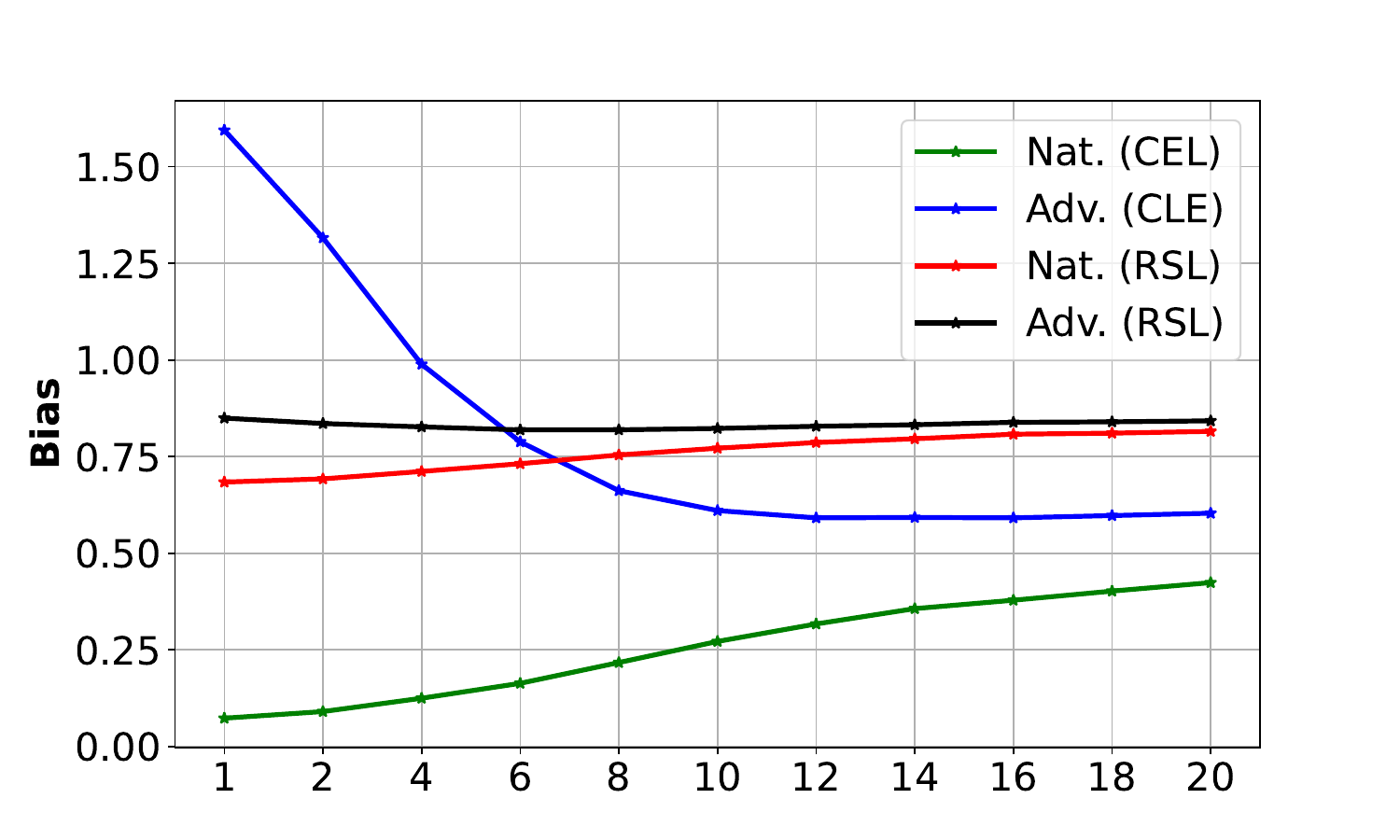}} &
          (e)  \makecell[c]{\includegraphics[width=0.3\linewidth]{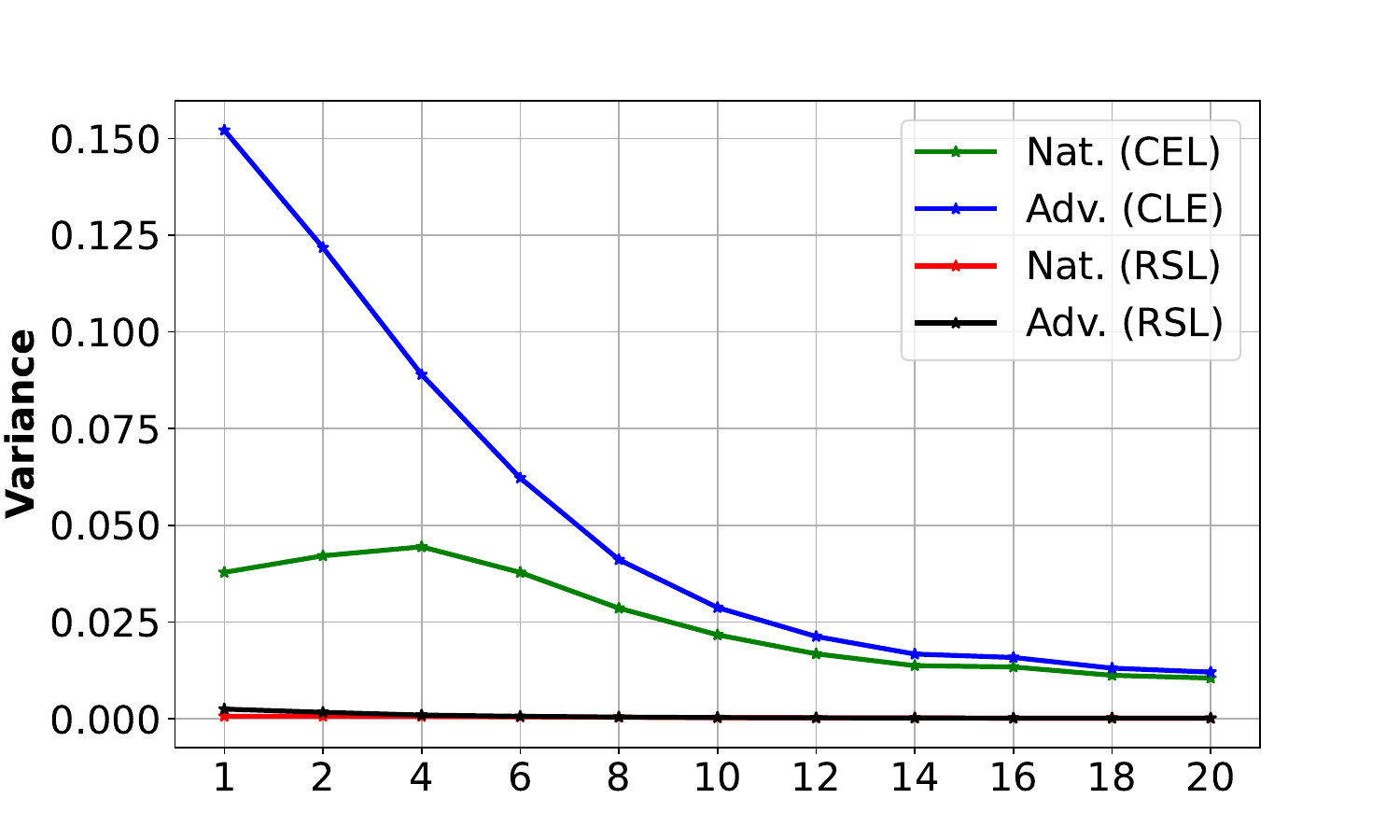}} &
          (f)\makecell[c]{\includegraphics[width=0.3\linewidth]{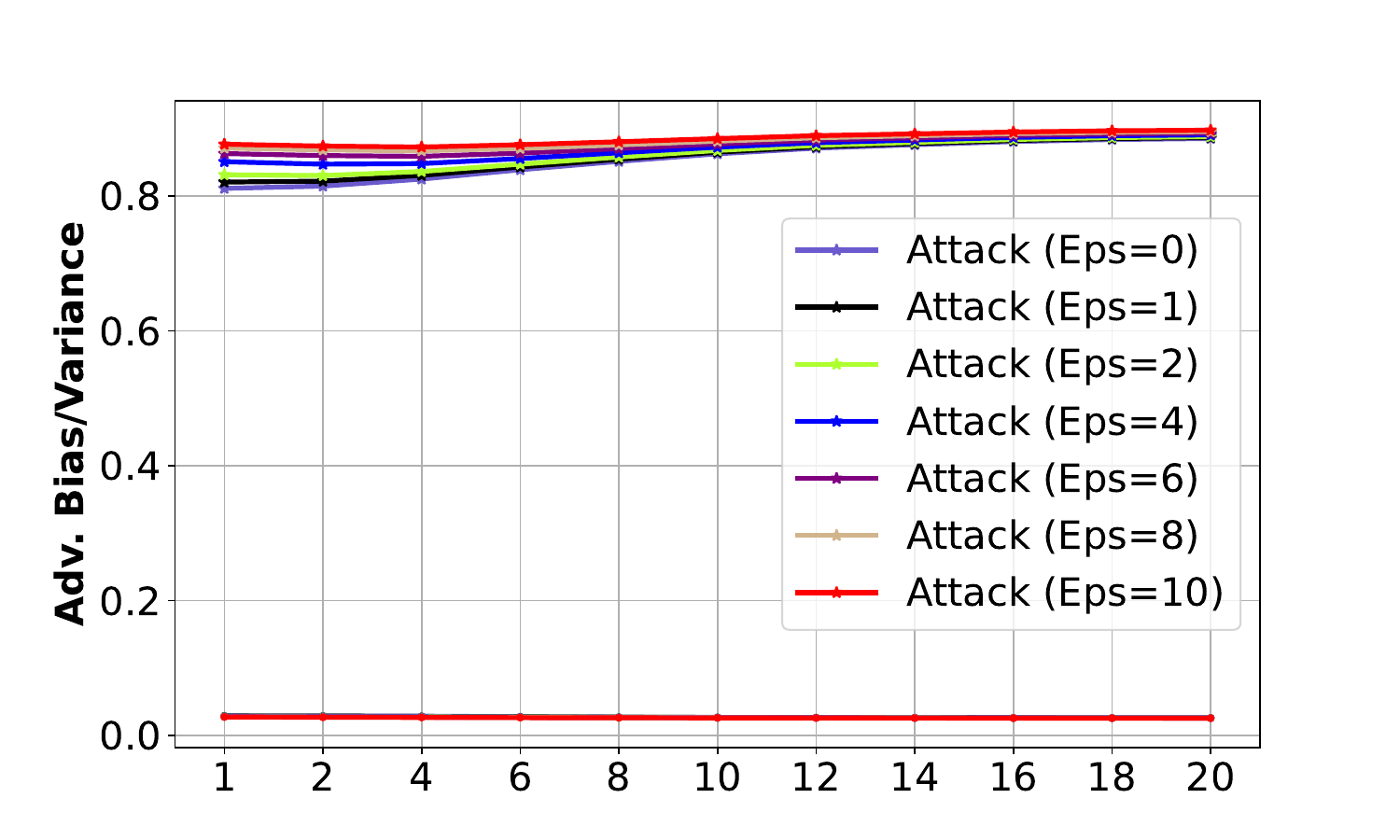}}\\
             
              % {(a) Training Adversarial Loss} & {(b) Test Accuracy} & {(c) Test Adversarial Accuracy} \\
       \end{tabular} %\vspace{-0.5em}
\caption{Bias, variance, and risk for $l_\infty$ adversarially trained models on the CIFAR10 using ResNet18. \label{BV}}
\end{figure*}
We begin our analysis by exploiting well-known bias-variance decomposition. 
Following \cite{yu2021understanding}, the adversarial expected risk can be decomposed as:
\begin{align}
& \mathbb{E}_{x,y}\mathbb{E}_{\mathcal{T}}\big[\|y-f(x,\mathcal{T})\| \big] \nonumber \\
      & = \mathbb{E}_{x,y}\big[ \|y-\hat{f}(x)\| \big] + \mathbb{E}_x\mathbb{E}_{\mathcal{T}}\big[ \|f(x,\mathcal{T})-\hat{f}(x)\| \big],
\end{align}
where $\hat{f}(x)=\mathbb{E}_{\mathcal{T}}f(x,\mathcal{T})$, and $\mathbb{E}_{\mathcal{T}}\big[\|y-f(x,\mathcal{T})\| \big]$ measures the average prediction error over different realizations of the training samples.
As a result, we uniformly separate the whole training data into two parts $\{\mathcal{T}_1,\mathcal{T}_2\}$, and use the training protocol highlighted in our SimpleAT.
The results are shown in Fig. \ref{BV}.
Here we refer to the bias and variance of natural images as ``natural bias and variance,'' while those of adversarial examples are referred to as ``adversarial bias and variance''

We investigate the impact of one cyclic learning rate and erasing-based data augmentation, as shown in Fig. \ref{BV} (a) and (b). 
Our findings reveal that natural variance dominates natural empirical risk, but adversarial bias remains the primary factor affecting adversarial empirical risk since adversarial variance is low. 
These two figures also show that data augmentation can decrease both natural and adversarial variances, leading to robustness and accuracy improvement.
Moreover, our observations confirm previous claims in \cite{smith2017cyclical} and \cite{chen2020group} because they mentioned that OCL or data augmentation can reduce the variance of a trained model, respectively. 
Interestingly, \cite{yu2021understanding} claimed that ``adversarial training increases the variance of the model predictions and thus leads to overfitting'', which verifies our result, \ie, combing RE and OCL makes the robust model achieves lower variance while mitigating robust overfitting.

We compare the difference between CEL and RSL. See Fig. \ref{BV} (c-e). 
We train our robust model under different perturbation budgets, \eg, $\epsilon$ is from 1 to 20. All the adversarial results are evaluated under the PGD-attack with 10 steps and $\epsilon=8$.
Fig. \ref{BV} (c) shows that RSL-trained models consistently achieve superior adversarial accuracy compared to CEL-trained models, albeit with slightly lower natural accuracy.

We compare the results of natural and adversarial bias/variance. See Fig. \ref{BV} (d) and (e). 
For CEL-trained models, the natural bias increases monotonically, and the adversarial bias decreases rapidly under a small perturbation budget and then flattens out.
The natural variance curve of the CEL-trained model is unimodal which is similar to the phenomenon observed by \cite{yu2021understanding}, and the adversarial variance curve is similar to the corresponding natural bias curve. 
On the other hand, RSL-trained models have a very different tendency for curves. 
Where both natural and adversarial bias change a little with changing the perturbation budgets, and two variance scores of RSL-trained models are extremely low. 
% These observations yield intriguing findings. 
Moreover, with small perturbation budgets, both adversarial bias and variance of CEL-trained models are totally higher. 
The gaps between natural and adversarial bias/variance are also large. This means the models are easily attacked. 
Then, with the increasing perturbation budget, the gaps become smaller, and the adversarial bias and variance are also becoming small enough. It means that the model becomes robust against adversarial attacks.

The aforementioned observations reveal intriguing findings.
Specifically, with small perturbation budgets, models trained with CEL exhibit significantly higher adversarial bias and variance. The gap between natural and adversarial bias/variance is substantial, indicating that these models are highly vulnerable to attacks. 
However, as the magnitude of perturbation increases, the gaps gradually narrow and both adversarial bias and variance decrease correspondingly. This suggests that the models become more robust against adversarial attacks.
Moreover, for a model trained with RSL, the difference in bias and variance between natural and adversarial examples is minimal, accompanied by smaller variances. 
It can be inferred that \textit{a robust model should exhibit a small gap between natural and adversarial bias while maintaining lower variances.}

Finally, the results for SimpleAT are summarized in Fig. \ref{BV} (f), which shows similar tendency curves with varying perturbation size $\epsilon$. This observation further supports our inference, which indicates that our SimpleAT indeed achieves good robustness and accuracy. 
Furthermore, we observe that the choice of loss function impacts the final bias and variance as shown in Fig. \ref{BV}.
We believe that the current methods of adversarial attack optimize a loss function with a formulation similar to that of the adversarial bias used for learning perturbation noise. 
It is worth noting that the rescaled square loss function is closely related to the definition of adversarial bias.
As a result, we revisit the rescaled square loss function and provide a more in-depth analysis in the next subsection.

\subsection{Relations to Logit Penalty Methods \label{sec42}}

We ﬁnally provide an in-depth analysis based on the RSL, by introducing a perspective as:

\noindent\textbf{Perspective 1.} RSL adaptively constrains logit features to be small in $L_2$ ball. 

We rewrite the formulation of RSL as follows:
\begin{equation}\label{eq10}
      \|\vk\cdot(\vf(\vx)-M\cdot \vy_{hot})\|_2^2 \approx \va\|\vf(\vx)\|^2_2-2\vb\vf^T(\vx)\vy_{hot},
\end{equation}
where $\va=\|\vk\|$ and $\vb = 2M\|\vk\|$. These two parameters remain constant when fixed during training.
In fact, RSL comprises two distinct functions: 
(i) the first one is to minimize the $L_2$-norm of the logit feature, \ie,~ $\vf(\vx)$; (ii) the second one is equal to the function of CEL (without the softmax operation).
Therefore, we argue that the primary distinction from CEL lies within the first item and \textit{the effect of RSL is due to its adaptive constraint on the L2-norm of logit features.}

Recent works in \cite{stutz2020confidence,wei2022mitigating,Liu2023MitigatingRO} have shown that overconfidence usually leads to overfitting or worse generalization. 
Therefore, recent work usually considers using logit penalty, logit normalization, and label smoothing. 
Logit penalty (LP) \cite{kornblith2021better} is to impose a penalty on the logit features, which can be formulated as $L_{LP} = L(\vx,\vy) + \beta \|\vf(\vx)\|_2^2$, where $\beta$ is the hyperparameter. 
Logit normalization (LN) \cite{wei2022mitigating} is to replace the original logit features with its $L_2$ normalization features, \ie, $\hat{\vf}=\vf/(\tau\|\vf\|)_2^2$. 
Furthermore, label smoothing (LS) has been extensively employed in prior research, which has indeed facilitated improvements in results as demonstrated by \cite{pang2021bag}.

\begin{figure*}[!t]
       \centering
       \footnotesize
       \setlength\tabcolsep{4pt}
       \begin{tabular}{ccc}
              (a) \makecell[c]{\includegraphics[width=0.28\linewidth]{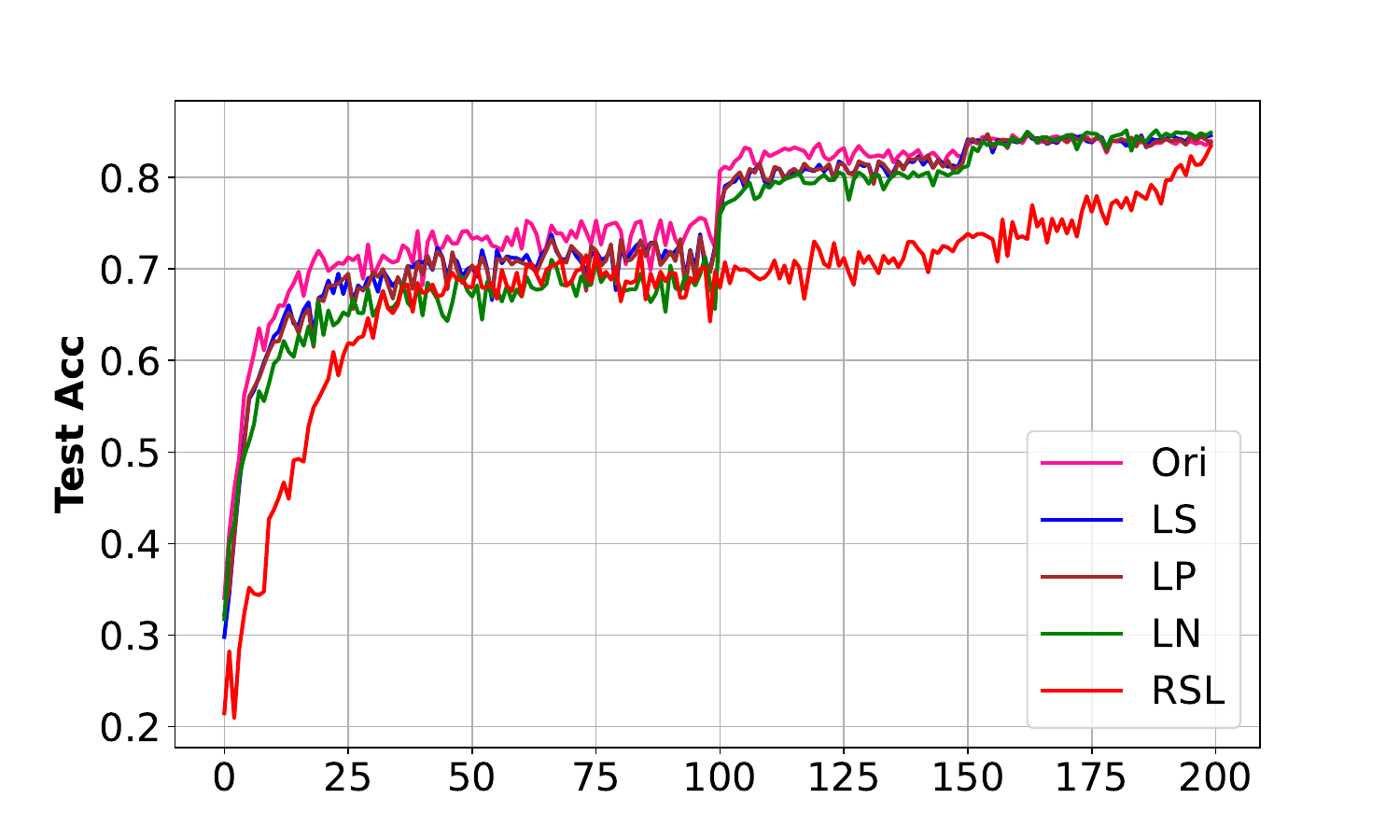}} &  
              (b) \makecell[c]{\includegraphics[width=0.28\linewidth]{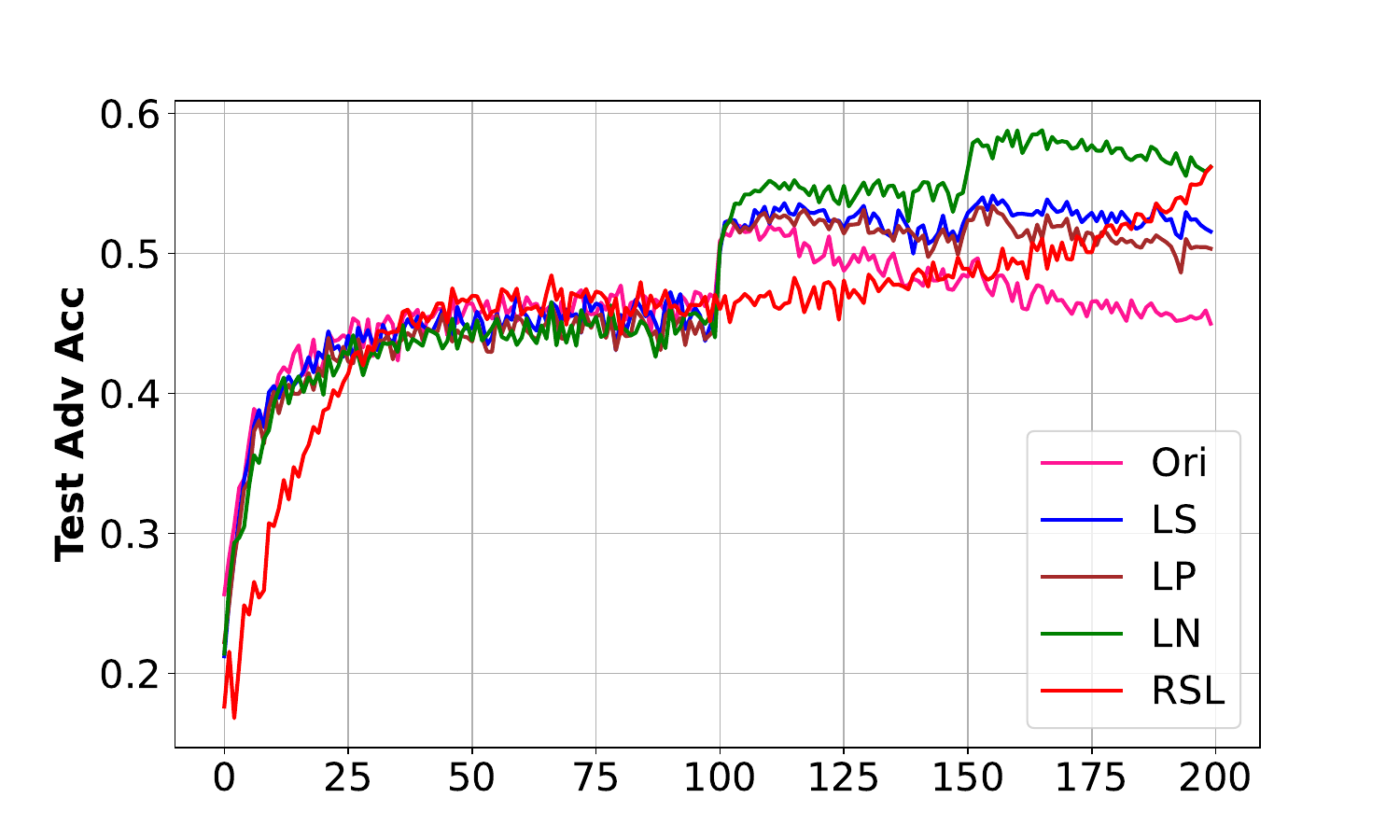}} & 
             \begin{tabular}{lccccc}
Method & Natural & PGD & AA  \\\midrule[1pt]
LS & 82.69 & 55.26 & 49.75\\
LP & 80.41 & 54.11 & 50.24  \\
LN & 80.23  & 58.80 & 46.38  \\
SimpleAT & 83.44  & 56.15 & 50.59  \\
\multicolumn{4}{c} {\small (c) Quantitive results.}\\
              \end{tabular}\\
       \end{tabular}\\
       \caption{Compared results of logit penalty methods trained on CIFAR-10. The backbone is ResNet-18. Subfigures (a) and (b) show the test accuracy curves and the final table reports the quantitive results. \label{logit}}
\end{figure*}
Based on these, we further conduct experiments on CIFAR-10 with ResNet-18. The results are shown in Fig. \ref{logit}.
We find that all logit penalty methods can mitigate overfitting problems and also achieve good results improvements. These results echo previous observations in \cite{stutz2020confidence,wei2022mitigating,Liu2023MitigatingRO}. That is, to achieve a good robust model, the conﬁdence level needs to be suppressed. 
LN achieves impressive robust accuracy against PGD-attack but remains vulnerable to strong AutoAttack. 
We believe that adjusting the hyperparameters such as $\beta$ or $\tau$ in logit penalty methods is challenging. 
While SimpleAT also has two additional parameters ($\vk$ and $M$),  they are easier to search for compared to finding the optimal hyperparameter of the regularization scheme.
In general, SimpleAT can achieve the overall best performance, due to its implicit logit constraints.

\section{Conclusion \label{sec8}}
We demonstrate how training protocols have a significant impact on model robustness. When controlling for various factors, we find that simple modifications to the prior training protocol can enhance model robustness. To this end, we introduce SimpleAT, a simple yet effective method that achieves a good trade-off between accuracy and robustness while mitigating robust overfitting. Our experiments have produced competitive results across various datasets, and our analysis also provides insights into the rationale behind these modifications. In the future, we aim to expand our analysis to include theoretical certifications of rescaled square loss and others.

\bibliographystyle{IEEEtran}
\bibliography{ref}

%\newpage

\begin{IEEEbiographynophoto}{Hong Liu}
%[{\includegraphics[width=1in,height=1.25in,clip,keepaspectratio]{liu-hong.jpg}}]{Hong Liu}
received his Ph.D. degree in computer science from Xiamen University. He is currently a JSPS Fellowship researcher working at the National Institute of Informatics, Japan. 
He was awarded the Japan Society for the Promotion of Science (JSPS) International Fellowship, the Outstanding Doctoral Dissertation Awards of both the China Society of Image and Graphics (CSIG) and Fujian Province, the Top-100 Chinese New Stars in Artificial Intelligence by Baidu Scholar, and the Notable Reviewer at ICLR 2023. He was a guest editor of IJCV and Electronics.
His research interests include trustworthy AI and deep learning theory.
\end{IEEEbiographynophoto}
%\end{IEEEbiography}
% your own custom macro containing the $\backslash${\tt{includegraphics}} command to make things
%\bf{If you include a photo:}\vspace{-33pt}
%\begin{IEEEbiography}[{\includegraphics[width=1in,height=1.25in,clip,keepaspectratio]{fig1}}]{Michael Shell}
%Use $\backslash${\tt{begin\{IEEEbiography\}}} and then for the 1st argument use $\backslash${\tt{includegraphics}} to declare and link the author photo.
%Use the author name as the 3rd argument followed by the biography text.
%\end{IEEEbiography}

\vfill

\end{document}